%% file: main.tex
\documentclass[letterpaper]{article} 
\usepackage[preprint]{aaai2027}  
\usepackage[hyphens]{url}  
\usepackage{graphicx} 
\usepackage{natbib}  
\usepackage{caption} 
\usepackage{algorithm}
\usepackage{algorithmic}

\usepackage{newfloat}
\usepackage{listings}
\DeclareCaptionStyle{ruled}{labelfont=normalfont,labelsep=colon,strut=off} 
\floatstyle{ruled}
\newfloat{listing}{tb}{lst}{}
\floatname{listing}{Listing}

\usepackage{booktabs}
\usepackage{multirow}
\usepackage{amsmath}
\usepackage{amssymb}

\title{OmniGround: Strengthening Omni-Modal Temporal Grounding via Attention-Consistency Training}
\author{
    Zixuan Chen\textsuperscript{\rm 1},\quad
    Depeng Wang\textsuperscript{\rm 2},\quad
    Hao Lin\textsuperscript{\rm 3},\quad
    Ke Xu\textsuperscript{\rm 1},\\
    Ya Guo\textsuperscript{\rm 2},\quad
    Huijia Zhu\textsuperscript{\rm 2}\corresponding,\quad
    Tanfeng Sun\textsuperscript{\rm 1},\quad
    Xinghao Jiang\textsuperscript{\rm 1}\corresponding
}
\affiliations{
    \textsuperscript{\rm 1}Shanghai Jiao Tong University\quad
    \textsuperscript{\rm 2}Ant Group\quad
    \textsuperscript{\rm 3}Independent Researcher
}

\begin{document}

\maketitle

\input{sections/abstract}


\input{sections/introduction}
\input{sections/related_work}

\input{sections/preliminary}
\input{sections/method}
\input{sections/experiments}
\input{sections/ablation}

\input{sections/conclusion}

\bibliography{references}

\input{sections/appendix}


\end{document}

%% file: sections/abstract.tex
\begin{abstract}
Omni-modal large language models can jointly understand video, audio, and text, yet they remain weak at \textbf{temporal grounding}, a capability essential for real-world audio-visual applications. To understand this limitation, we look beyond final predictions and examine the model's internal temporal evidence. We identify a \emph{temporal generation gap}: the internal attention of the model localizes the target moment more accurately than the timestamps it generates. To close this gap, we propose \textbf{TACG} (Temporal Attention-Consistency Grounding), a post-training method that treats internal attention as a reliable temporal pivot and aligns it with both the model prediction and the ground-truth annotation through attention-consistency training. We further build a 580K-sample training dataset by reconstructing temporal QA from dense audio-visual annotations and organizing general omni-modal instruction data. Using TACG and this training dataset, we train \textbf{OmniGround}, a general-purpose omni-modal model with stronger temporal grounding capability, without injecting any additional temporal cues. OmniGround reaches state-of-the-art omni-modal temporal grounding, including 37.5 mIoU on LongVALE and 35.7 mIoU on UnAV-100, while staying competitive on both general audio-visual understanding and vision-only grounding.
\end{abstract}

%% file: sections/introduction.tex
\section{Introduction}
\label{sec:introduction}

\begin{figure*}
    \centering
    \includegraphics[width=\linewidth]{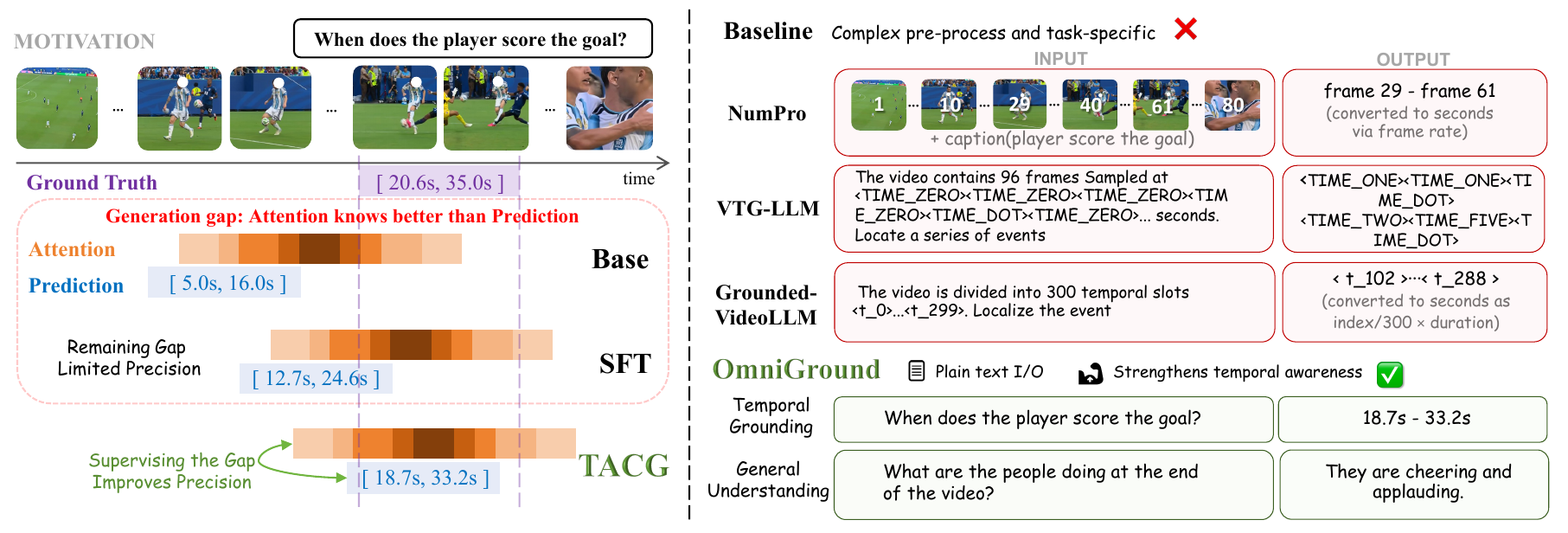}
    \caption{\textbf{Overview of OmniGround.} \textbf{Left:} both the base model and the stronger SFT model exhibit a \emph{temporal generation gap}: internal attention localizes the target event more accurately than the generated timestamp. TACG closes it to achieve stronger grounding. \textbf{Right:} unlike methods that rely on additional temporal inputs or specialized output formats, OmniGround strengthens temporal grounding while preserving a general-purpose omni-modal interface.}
    \label{fig:intro}
\end{figure*}
Omni-modal large language models~\cite{qwen25omni,ola,nextgpt}, which jointly perceive video, audio, and text within a single backbone, have advanced rapidly and now excel at a wide range of multimodal understanding tasks. As these models move toward real-world audio-visual applications such as video editing, content moderation, and multimodal retrieval, recognizing \emph{what} happens no longer suffices: a model must also pinpoint \emph{when} it happens. This temporal capability is critical for practical audio-visual applications, with temporal video grounding (TVG) serving as a central benchmark. Yet current omni-modal models still perform poorly on TVG despite their strong general understanding.

Existing temporal grounding methods mainly follow two directions. One family modifies the model or its inputs by injecting explicit temporal information, such as special temporal tokens, frame markers, or task-specific heads~\cite{vtgllm,numpro,edvtg}. Although effective for localization, these designs make temporal grounding depend on customized input formats and specialized output parsing, which limits their compatibility with a general-purpose omni-modal interface. A second family improves grounding purely through post-training, such as reinforcement learning~\cite{timer1} or distillation from a stronger teacher~\cite{videoopd}. This training-only direction preserves the inference interface and better maintains general ability, but does not directly exploit the model's own temporal evidence. We adopt this philosophy, but push it further.

We begin with a diagnostic question: when an omni-modal model grounds imprecisely, does it fail to \emph{localize} the event, or merely to \emph{express} the time? For each query, we compare the interval read from the model's internal attention with its generated timestamp, and measure which lands closer to the ground truth. Across multiple omni-modal models and grounding datasets, the attention-derived interval is more accurate than the prediction on a clear majority of samples, up to 78.3\%. We call this the \textit{temporal generation gap}: the model knows where to look but cannot accurately say when. This pattern holds for both base models and models with stronger grounding ability, indicating that the gap is not merely a weak-model artifact, as illustrated in Figure~\ref{fig:intro}. If left unaddressed, this gap limits further improvements in temporal localization precision.

Building on this diagnosis, we propose \textbf{Temporal Attention-Consistency Grounding} (TACG). Its core term aligns the generated timestamp with the model's own internal attention, using the more reliable internal signal to pull the prediction onto the target and directly close the temporal generation gap. Two auxiliary terms anchor this correction to the ground truth: one supervises the attention directly against the annotation, making the attention itself localize the target more accurately; the other supervises the prediction by its temporal distance to the annotation, supplying the distance-aware gradient that cross-entropy lacks. TACG achieves stronger temporal grounding than standard supervised fine-tuning on the same data, without introducing additional temporal inputs or model parameters.

We also build a 580k-sample omni-modal training dataset that supports both temporal grounding and general understanding. Starting from the dense audio-visual annotations in AVID~\cite{avid} and LongVALE~\cite{longvale}, we extract high-quality temporal segments and reformulate them into grounding, dense captioning, and temporal question-answering tasks. We further incorporate general omni-modal instruction data from several sources~\cite{llavavideo,ola,avqa,fortisavqa} to preserve broad audio-visual understanding. Using this dataset and TACG, we train \textbf{OmniGround} from Qwen2.5-Omni~\cite{qwen25omni}. As shown in Figure~\ref{fig:intro}, OmniGround improves temporal grounding without the extra temporal inputs or specialized output formats used by prior baselines. We evaluate OmniGround on seven benchmarks, including three omni-modal grounding benchmarks, two vision-only grounding benchmarks, and two general omni-modal understanding benchmarks. OmniGround sets a new state of the art on omni-modal temporal grounding while staying competitive on the other two categories. Our contributions are summarized as follows:

\begin{itemize}
\item We identify the \textit{temporal generation gap} in omni-modal language models and propose TACG, a post-training objective that aligns the model's prediction with its internal attention, consistently improving over standard supervised fine-tuning under identical training data.
\item We build a 580k omni-modal temporal-awareness dataset by reformulating dense audio-visual annotations into temporal grounding and question-answering tasks, applying quality filtering, and balancing the resulting data with curated general data. We train OmniGround on this dataset.
\item \textbf{OmniGround} reaches state-of-the-art omni-modal temporal grounding and stays competitive on general omni-modal understanding and vision-only grounding. We will release the dataset, model, and training framework.
\end{itemize}

%% file: sections/related_work.tex
\section{Related Work}
\label{sec:related_work}

\noindent\textbf{Omni-Modal Language Models.}
Recent work extends large language models beyond text to perceive multiple modalities. Early video LLMs such as VideoChat~\cite{videochat}, Video-ChatGPT~\cite{videochatgpt}, and VideoLLaMA~\cite{videollama} couple a visual encoder with an LLM, while PandaGPT~\cite{pandagpt} and NExT-GPT~\cite{nextgpt} add audio. More recent omni-modal models further unify audio, vision, and text within a shared modeling framework, enabling general audio-visual-text understanding~\cite{ola,qwen25omni}. While these models have achieved strong general audio-visual understanding, they are not explicitly optimized for precise temporal grounding. As a result, they often recognize the relevant event but fail to generate accurate temporal boundaries. Our work strengthens temporal grounding in a general-purpose omni-modal model.

\noindent\textbf{Model Mechanism Analysis.}
Prior studies show that model behavior can be analyzed through internal signals~\cite{zhao2024first,Palit_2023_ICCV}. Inspired by this perspective, we inspect internal attention in omni-modal temporal grounding.

\noindent\textbf{Temporal Video Grounding.}
Temporal video grounding (TVG) localizes the moment described by a natural language query, and existing methods differ in how they acquire this ability. One family injects explicit temporal machinery---regression heads in specialist architectures~\cite{2dtan,univtg,eatr,snag}, or, in video LLMs, time-aware encoders~\cite{timechat}, special temporal tokens~\cite{vtgllm,groundedvideollm,chronusomni}, frame-index markers~\cite{numpro}, and localization heads~\cite{edvtg,vtimellm,hawkeye,timesuite,trace,videochatflash,chatvtg,revisionllm}. These components depart from the standard backbone, add inference cost, and often weaken general question answering. A second family leaves the architecture and inference untouched and strengthens temporal grounding through additional post-training stages, such as reinforcement learning with IoU rewards~\cite{timer1} or distillation from a stronger teacher~\cite{videoopd}. Our method instead strengthens fine-tuning itself, turning the model's internal attention into a self-supervision signal.

%% file: sections/preliminary.tex
\section{Preliminary: Temporal Generation Gap}
\label{sec:preliminary}
\begin{figure}[t]
\centering
\includegraphics[width=0.98\linewidth]{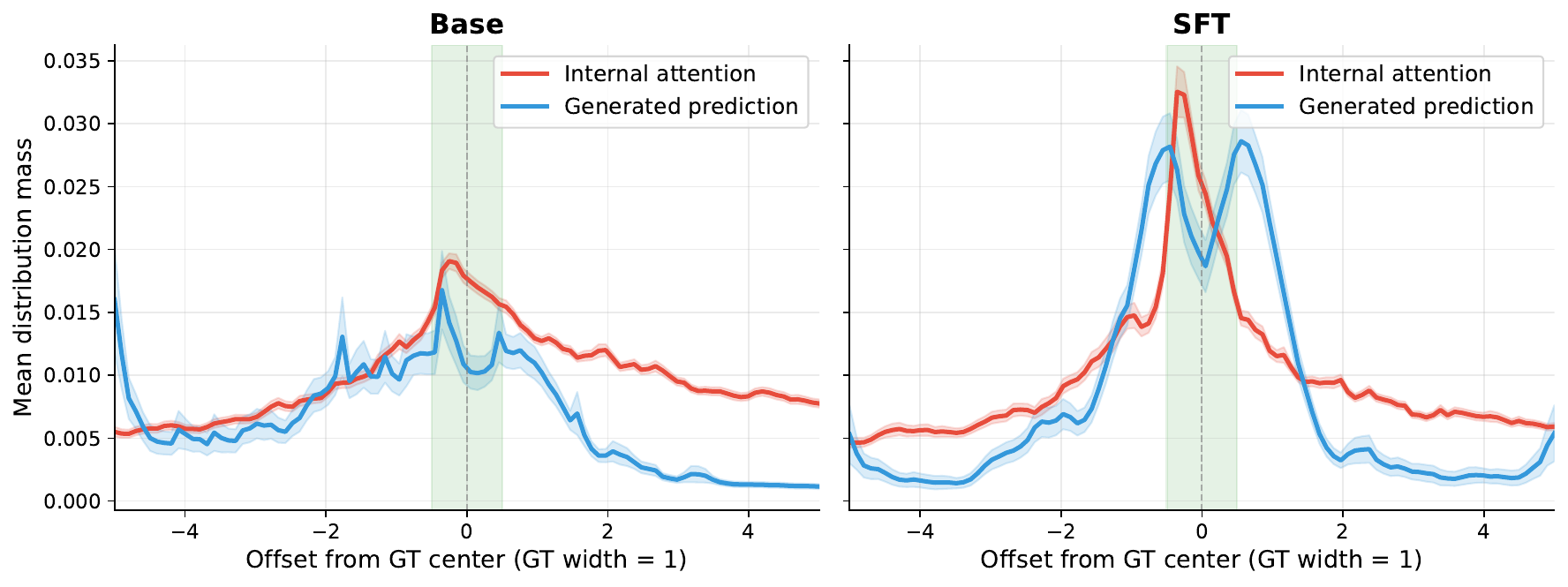}\\
\includegraphics[width=0.98\linewidth]{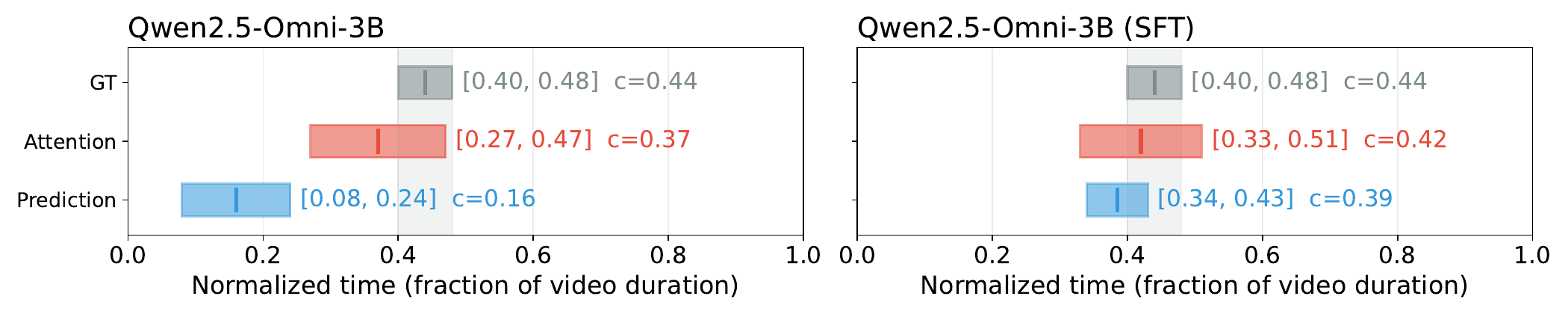}\\[2pt]
\makebox[0.49\linewidth]{\small (a) Qwen2.5-Omni-3B (base)}%
\makebox[0.49\linewidth]{\small (b) Qwen2.5-Omni-3B (SFT)}
\caption{\textbf{Visualization of the temporal generation gap.} \emph{Top:} the attention averaged over samples and aligned to the ground-truth event (green), overlaid with the averaged prediction. \emph{Bottom:} the corresponding averaged attention and predicted intervals. In both base and SFT models, the attention localizes the event more accurately than the prediction.}
\label{fig:gap_vis}
\end{figure}

\begin{table}[t]
\centering
\small
\setlength{\tabcolsep}{2pt}
\begin{tabular}{lcccc}
\toprule
 & \multicolumn{2}{c}{AVID} & \multicolumn{2}{c}{LongVALE} \\
\cmidrule(lr){2-3}\cmidrule(lr){4-5}
Model & Attn.\ wins$^\dagger$ & mIoU & Attn.\ wins$^\dagger$ & mIoU \\
\midrule
Qwen2.5-Omni-3B & $73.9\%$ & $7.4$ & $78.3\%$ & $6.3$ \\
Qwen2.5-Omni-3B (SFT) & $57.7\%$ & $17.9$ & $60.0\%$ & $20.9$ \\
Qwen2.5-Omni-7B & $75.7\%$ & $6.2$ & $72.1\%$ & $12.1$\\ 
Qwen3-Omni & $65.0\%$ & $14.1$ & $69.7\%$ & $10.6$ \\
OLA & $73.0\%$ & $10.3$ & $70.3\%$ & $7.7$ \\
\bottomrule
\end{tabular}
\caption{\textbf{The temporal generation gap across models and datasets.} For each benchmark, all statistics are computed on a fixed evaluation subset of 500 examples randomly sampled from its official test split. \emph{Attn.\ wins}$^\dagger$ reports the fraction of non-tied samples where $\mathrm{IoU}(I_{\text{att}}, I_{\text{gt}}) > \mathrm{IoU}(I_{\text{pred}}, I_{\text{gt}})$. \emph{mIoU} is the grounding accuracy of the generated interval. This gap is consistent across models and datasets.}
\label{tab:js_gap}
\end{table}

We validate the temporal generation gap across multiple omni-modal models on AVID and LongVALE, including both base models and models with stronger grounding ability. For each benchmark, we construct the diagnostic evaluation split by randomly sampling 500 examples from its official test split and use this fixed subset for all preliminary-analysis statistics. For each sample we obtain three temporal intervals and compare them against the ground truth by temporal IoU. The ground-truth interval $I_{\text{gt}}$ and the generated interval $I_{\text{pred}}$ are taken directly. We extract the attention interval $I_{\text{att}}$ from the attention of the last prefill token to the audio-visual tokens at a mid-to-late layer of the model. This attention is aggregated over spatial positions into a temporal distribution, from which we select the shortest contiguous window covering $50\%$ of the cumulative mass. This gives three intervals for each case, $I_{\text{gt}}$, $I_{\text{pred}}$, and $I_{\text{att}}$; we compare $\mathrm{IoU}(I_{\text{att}}, I_{\text{gt}})$ with $\mathrm{IoU}(I_{\text{pred}}, I_{\text{gt}})$ to determine whether the attention interval or the generated prediction is more accurate, and use the same intervals for visualization. For Qwen2.5-Omni-3B, we use layer~27; additional details, including layer choices for other models, layer sensitivity, cumulative-mass threshold sensitivity, and the SFT setup, are provided in the Appendix.

Figure~\ref{fig:gap_vis} visualizes the internal attention concentrating on the target region while the generated interval is displaced, and Table~\ref{tab:js_gap} shows that $I_{\text{att}}$ overlaps the ground truth better than $I_{\text{pred}}$ on most valid samples across five omni-modal models and two datasets, despite large differences in grounding mIoU. Together, these results establish that the temporal generation gap is consistent across models and datasets. After SFT, both the attention interval and the generated prediction become more accurate, yet the instance-level gap remains, suggesting that unresolved attention--prediction misalignment limits further improvements in temporal localization precision. This motivates directly aligning attention, prediction, and ground truth in the training objective introduced next.

%% file: sections/method.tex
\section{Method: OmniGround}
\label{sec:method}
\begin{figure*}[t]
\centering
\includegraphics[width=\textwidth]{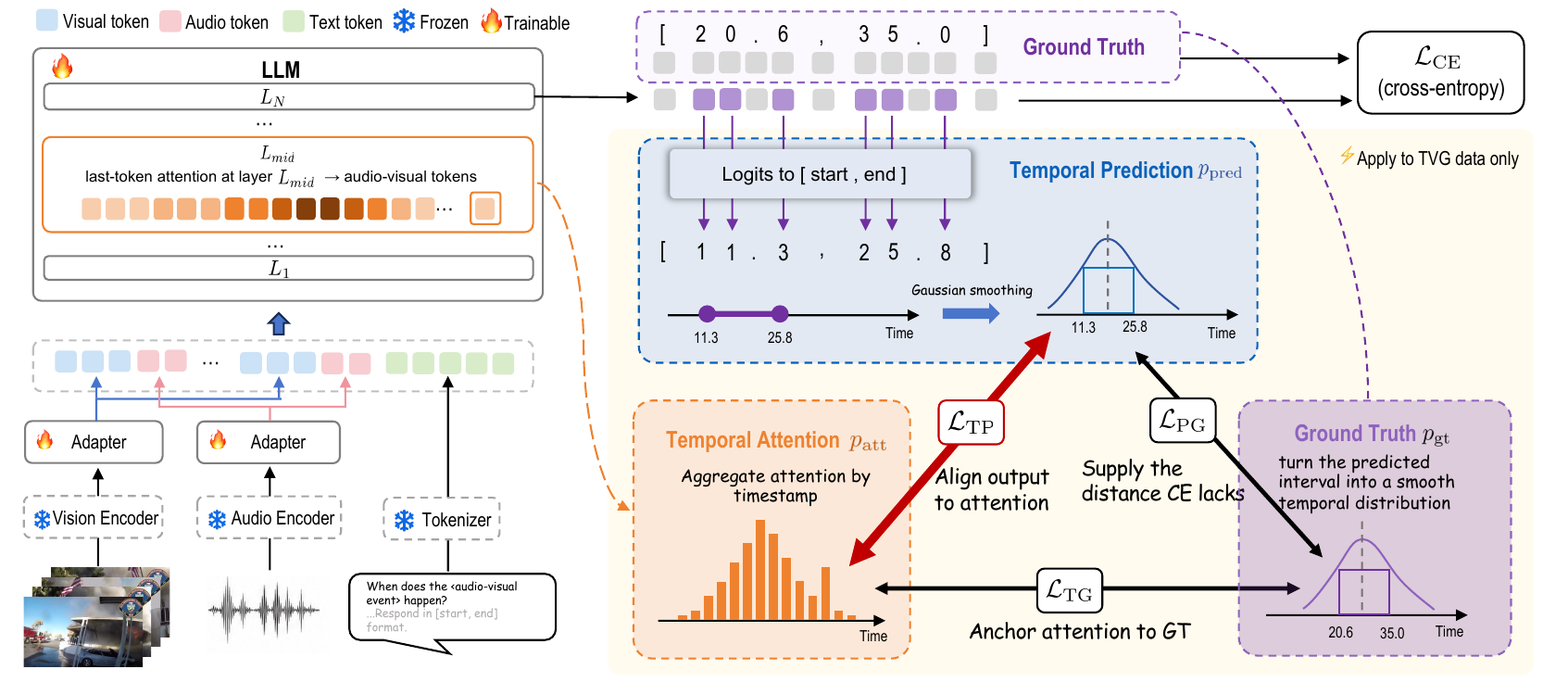}
\caption{Overall architecture and TACG training of OmniGround.}
\label{fig:framework}
\end{figure*}

We present OmniGround, an omni-modal temporal grounding model built on Qwen2.5-Omni. We train it with TACG on our 580k-sample omni-modal training dataset. The framework is illustrated in Figure~\ref{fig:framework}.

\subsection{Temporal Attention-Consistency Grounding}
\label{sec:tccg}

The preliminary analysis motivates supervising the generated timestamp with the model's internal attention while anchoring both signals to the ground truth. Although temporal IoU measures their agreement at evaluation time, it is unsuitable as a \emph{training} signal: it is non-differentiable and least informative exactly when most needed. Once two intervals are disjoint, its gradient vanishes everywhere, giving no indication of which direction, or how far, to move the prediction---precisely the regime that dominates early training. We therefore recast attention, prediction, and ground truth as continuous temporal distributions over the $T$ time steps and enforce their mutual consistency with divergences that stay informative under non-overlap (Jensen--Shannon and Wasserstein-1, chosen per edge below).

\textbf{Attention--prediction consistency: the core objective.}
The central constraint of TACG forces the generated timestamp to agree with the localization already present in the internal attention. We implement this constraint by expressing both quantities as temporal distributions over the $T$ time steps.

We first extract the implicit localization signal from the attention at a selected intermediate layer $l$. We compute the attention from the last prefill token to all media tokens (both video and audio) and aggregate the mass that falls on each of the $T$ interleaved time steps:
\begin{equation}
p_{\text{att}}[t] = \sum_{i \in \mathcal{C}_t} \frac{\exp(q^\top k_i / \sqrt{d})}{\sum_{j} \exp(q^\top k_j / \sqrt{d})},
\label{eq:tad}
\end{equation}
where $q$ is the query of the last token, $k_i$ ranges over all video and audio tokens, and $\mathcal{C}_t$ is the set of media tokens at time step $t$. As the attention is normalized over all media tokens, $p_{\text{att}}$ is automatically a valid distribution over the $T$ steps, and gradients flow through the layer-$l$ query and key projections, allowing the consistency loss to refine the attention itself.

We then convert the predicted timestamp into a continuous temporal distribution. Since standard supervised fine-tuning never decodes the logits, we operate on them directly. At each digit position of the timestamp tokens we form the soft expectation over digit values, round it, apply a straight-through estimator so that gradients still reach the logits, and assemble the start and end times by place value:
\begin{equation}
\begin{aligned}
\hat{d}_i &= \sum_{d\in\mathcal{D}} d\cdot\text{softmax}(\text{logits}_i[\mathcal{D}])_d, \quad \bar{d}_i = \lfloor \hat{d}_i \rceil, \\
\tilde{d}_i &= \bar{d}_i - \text{sg}(\hat{d}_i) + \hat{d}_i, \\
\hat{t}_s &= \sum_{i} \tilde{d}^{\,s}_i\, 10^{p_i}, \quad \hat{t}_e = \sum_{i} \tilde{d}^{\,e}_i\, 10^{p_i},
\end{aligned}
\label{eq:round_method}
\end{equation}
where $\mathcal{D} = \{0, \ldots, 9\}$ are the digit token indices, $\lfloor \cdot \rceil$ rounds to the nearest integer, $\text{sg}(\cdot)$ is the stop-gradient (detach) operator, $p_i$ is the (signed) place value of digit $i$, and $\tilde{d}^{\,s}_i, \tilde{d}^{\,e}_i$ are the digits of the start and end timestamps. The second line is the straight-through estimator: the forward pass uses the rounded digit $\bar{d}_i$ (the two soft terms cancel), while the backward pass routes gradients through the soft expectation $\hat{d}_i$ to the logits. We map the reconstructed interval $[\hat{t}_s, \hat{t}_e]$, normalized by the video duration, to a continuous distribution $p_{\text{pred}}$ through a \emph{Gaussian projection}: a Gaussian centered at the interval center $\mu_{\hat t}$ with standard deviation $\sigma = 0.15 \cdot (\hat{t}_e - \hat{t}_s)$ proportional to the predicted width. This projection is a smoothing operator, it places the prediction in the same distributional space as the attention so the two can be compared directly, and gives the interval smooth support so that its divergence from other distributions varies continuously with the temporal offset rather than vanishing once they no longer overlap. 

With both distributions in place, the core objective is their Wasserstein-1 distance, $\mathcal{L}_{\text{tp}} = \mathcal{W}_1(p_{\text{att}}, p_{\text{pred}})$. This attention--prediction edge transfers the reliable localization already present in the internal attention onto the generated timestamps, directly closing the temporal generation gap. The divergence on each edge (Jensen--Shannon or Wasserstein-1) is chosen by ablation (Appendix).

\textbf{Ground-truth anchoring: two auxiliary constraints.}
The self-consistency edge alone does not tie either quantity to the true interval, and could in principle drive the two into mutual agreement away from it. Two auxiliary edges therefore anchor the transfer to the ground truth. We apply the same Gaussian projection, with identical parameters, to the ground-truth interval $[t_s^*, t_e^*]$, obtaining a distribution $p_{\text{gt}}$ that is detached from the computation graph and serves as the fixed supervision target. The attention--ground-truth edge keeps the internal signal aligned with the true segment, so that it remains a trustworthy target for the core edge. The prediction--ground-truth edge supplies the distance-aware supervision that cross-entropy lacks:
\begin{equation}
\mathcal{L}_{\text{tg}} = \mathrm{JS}(p_{\text{att}}, p_{\text{gt}}), \quad \mathcal{L}_{\text{pg}} = \mathcal{W}_1(p_{\text{pred}}, p_{\text{gt}}).
\label{eq:loss_tg}
\end{equation}

\textbf{Full objective.}
The three edges form a triangle of consistency constraints. With edge weights $\lambda_{\text{tp}}$, $\lambda_{\text{tg}}$, and $\lambda_{\text{pg}}$, the full training objective is
\begin{equation}
\mathcal{L}_{\text{total}} = \mathcal{L}_{\text{CE}}
+ \lambda_{\text{tp}}\, \mathcal{L}_{\text{tp}}
+ \lambda_{\text{tg}}\, \mathcal{L}_{\text{tg}}
+ \lambda_{\text{pg}}\, \mathcal{L}_{\text{pg}},
\label{eq:total_loss}
\end{equation}

\begin{figure}[t]
    \centering
    \includegraphics[width=\linewidth]{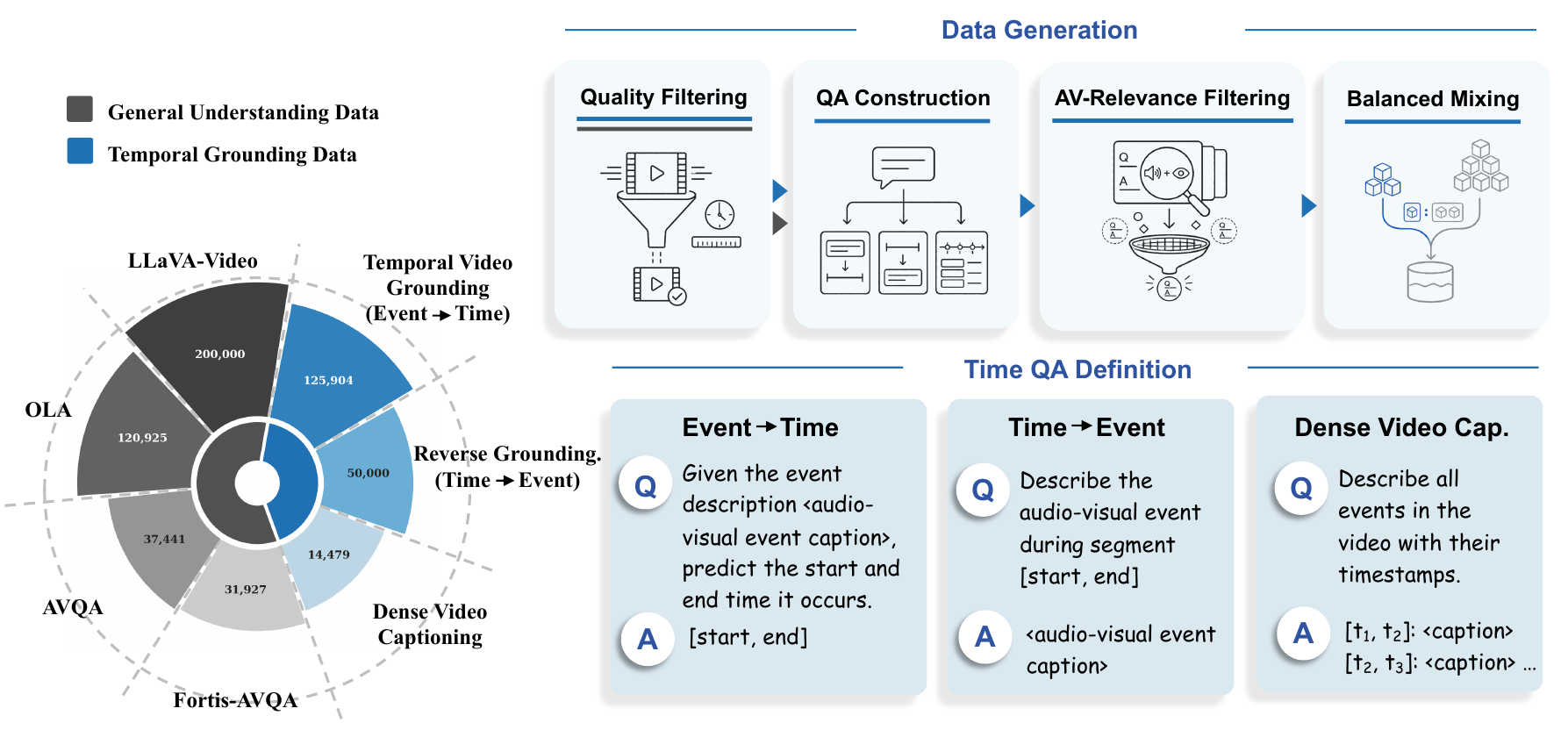}
  \caption{\textbf{Left:} dataset composition.
  \textbf{Right:} four-stage construction pipeline and representative QA templates for the three grounding formats.}
    \label{fig:data}
\end{figure}
\subsection{Data Construction and Training Strategy}
\label{sec:data}

After \emph{Quality Filtering} removes low-resolution videos and overly short clips (grounding videos under 30\,s and general clips under 2\,s), we assemble the general understanding data from four balanced sources: 121k LLaVA-Video~\cite{llavavideo} samples re-annotated by OLA~\cite{ola} from speech and visual perspectives, 37k AVQA~\cite{avqa} samples for scene sound, 32k FortisAVQA~\cite{fortisavqa} samples for music, and 200k original LLaVA-Video samples paired with audio for general video proficiency. For grounding data, we use only the \emph{training-split videos and dense temporal captions} of AVID~\cite{avid} and LongVALE~\cite{longvale}. We reformat the training captions into three complementary tasks---\emph{Temporal Video Grounding} (event~$\rightarrow$~time), \emph{Reverse Grounding} (time~$\rightarrow$~event), and \emph{Dense Video Captioning} (video~$\rightarrow$~timed events). We then apply \emph{Audio-Visual Relevance Filtering}, using Gemini~3.1~Pro to remove QA pairs whose text does not require audio-visual evidence. A final \emph{Balanced Mixing} step combines the two streams, yielding 190k grounding samples (126k TVG, 50k Reverse Grounding, and 14k DVC) and 390k general samples, for a 580k-sample dataset at a grounding-to-general ratio of roughly $1{:}2$; the composition and pipeline are summarized in Figure~\ref{fig:data} and detailed in the Appendix.

We apply TACG only to TVG samples because it needs an explicit predicted time interval to construct $p_{\text{pred}}$, which is available only when the target output is a start--end timestamp. All samples are trained jointly: TVG samples use the full objective $\mathcal{L}_{\text{total}}$, while Reverse Grounding, DVC, and general QA samples use cross-entropy $\mathcal{L}_{\text{CE}}$ only. Implementation details are provided in the Appendix.

%% file: sections/experiments.tex
\section{Experiments}
\label{sec:experiments}

\subsection{Experimental Setup}
All experiments are conducted on 32 NVIDIA A100 GPUs. Detailed configurations are provided in the Appendix.

\textbf{Evaluation Benchmarks.}
We evaluate on seven benchmarks: AVID~\cite{avid}, LongVALE~\cite{longvale}, and UnAV-100~\cite{geng2023dense} for omni-modal grounding, Charades-STA~\cite{charades} and ActivityNet-Captions~\cite{activitynet} for vision-only grounding, and WorldSense~\cite{worldsense} and Daily-Omni~\cite{dailyomni} for general omni-modal understanding. For UnAV-100, we convert its temporal annotations into QA pairs for large-model evaluation, with details provided in the Appendix.

\textbf{Baselines.}
We compare against 17 baselines in two groups:
\textit{vision-only models}, including VideoChat~\cite{videochat}, VTimeLLM~\cite{vtimellm}, TimeChat~\cite{timechat}, HawkEye~\cite{hawkeye}, VTG-LLM~\cite{vtgllm}, Grounded-VideoLLM~\cite{groundedvideollm}, TimeSuite~\cite{timesuite}, NumPro~\cite{numpro}, ED-VTG~\cite{edvtg}, and Time-R1~\cite{timer1};
and \textit{omni-modal models}, including VideoLLaMA2~\cite{videollama2}, OLA~\cite{ola}, Qwen2.5-Omni~\cite{qwen25omni}, Qwen3-Omni~\cite{qwen3omni}, LongVALE~\cite{longvale}, ChronusOmni~\cite{chronusomni}, and the closed-source Gemini~3.1~Pro~\cite{gemini}, which we report for reference only.

\subsection{Temporal Grounding Performance}

Table~\ref{tab:main} and Table~\ref{tab:unav100_tvg} report temporal grounding results across the five grounding benchmarks. We highlight three observations.

\textbf{Omni-modal temporal grounding.}
As shown in Table~\ref{tab:main}, OmniGround achieves the best open-model performance on the audio-visual AVID and LongVALE benchmarks. The 7B model reaches 65.9 mIoU on AVID and 37.5 mIoU on LongVALE, while the 3B model already surpasses all open baselines on AVID. Since our grounding data is reconstructed from the training-split source videos of AVID and LongVALE, we further evaluate on UnAV-100 to isolate out-of-domain audio-visual grounding performance. As shown in Table~\ref{tab:unav100_tvg}, OmniGround reaches 35.7 mIoU with the 7B model and outperforms evaluated open baselines at the stricter R@0.5 and R@0.7 thresholds. Together, these results show that OmniGround sets a new open-model state of the art in omni-modal temporal grounding.

\begin{table}[h]
\centering
\setlength{\tabcolsep}{6pt}
\small
\begin{tabular}{l|cccc}
\toprule
Method & R@0.3 & R@0.5 & R@0.7 & mIoU \\
\midrule
VideoLLaMA2 & 13.3 & 7.0 & 2.3 & 11.5 \\
OLA & 13.6 & 6.6 & 2.0 & 13.1 \\
Qwen2.5-Omni (3B) & 20.7 & 11.8 & 5.5 & 15.2 \\
Qwen2.5-Omni (7B) & 25.9 & 16.2 & 6.7 & 20.1 \\
Qwen3-Omni & 26.5 & 17.5 & 8.0 & 21.8 \\
LongVALE & 11.3 & 6.1 & 3.2 & 15.0 \\
ChronusOmni & \textbf{47.3} & \underline{33.8} & \underline{20.7} & \underline{34.1} \\
Gemini 3.1 Pro$^\star$ & 61.6 & 47.6 & 28.3 & 44.1 \\
\midrule
OmniGround(3B)$^\dagger$ & 36.0 & 21.6 & 10.9 & 25.9 \\
OmniGround(7B)$^\dagger$ & 42.3 & 32.7 & 19.1 & 32.2 \\
\textbf{OmniGround(3B)} & 40.1 & 24.0 & 13.3 & 28.9 \\
\textbf{OmniGround(7B)} & \underline{46.1} & \textbf{34.2} & \textbf{22.2} & \textbf{35.7} \\
\bottomrule
\end{tabular}
\caption{Evaluation on UnAV-100. $^\star$Closed-source, for reference only. $^\dagger$Trained on the same data but with standard SFT (cross-entropy).}
\label{tab:unav100_tvg}
\end{table}

\begin{table*}[t]
\centering
\setlength{\tabcolsep}{1pt}
\small
\begin{tabular}{l|cccc|cccc|cccc|cccc}
\toprule
\multirow{3}{*}{Method} & \multicolumn{8}{c|}{Omni-modal Grounding} & \multicolumn{8}{c}{Vision-only Grounding} \\
\cmidrule(lr){2-9} \cmidrule(lr){10-17}
& \multicolumn{4}{c|}{AVID-TVG} & \multicolumn{4}{c|}{LongVALE-TVG} & \multicolumn{4}{c|}{Charades-STA} & \multicolumn{4}{c}{ActivityNet-Captions} \\
& R@0.3 & R@0.5 & R@0.7 & mIoU & R@0.3 & R@0.5 & R@0.7 & mIoU & R@0.3 & R@0.5 & R@0.7 & mIoU & R@0.3 & R@0.5 & R@0.7 & mIoU \\
\midrule
\multicolumn{17}{l}{\textit{Vision-only Models}} \\
VideoChat & 7.6 & 5.2 & 2.5 & 5.2 & 2.2 & 0.9 & 0.4 & 3.0 & 32.8 & 8.6 & 0.0 & 25.9 & 23.5 & 12.6 & 6.0 & 17.4 \\
VTimeLLM & 9.2 & 4.1 & 1.0 & 7.4 & 9.1 & 4.3 & 1.9 & 7.9 & 52.0 & 27.7 & 11.1 & 31.5 & 44.0 & 27.8 & 14.3 & 30.4 \\
TimeChat & 9.8 & 6.9 & 3.0 & 7.8 & 7.2 & 3.4 & 1.6 & 6.4 & 47.8 & 31.6 & 14.6 & 31.6 & 30.2 & 16.9 & 8.2 & 21.8 \\
HawkEye & 7.5 & 4.1 & 1.7 & 5.2 & 15.3 & 7.8 & 2.8 & 10.9 & 50.6 & 31.4 & 14.5 & 33.7 & 49.1 & 29.3 & 10.7 & 32.7 \\
VTG-LLM & 9.0 & 6.4 & 2.9 & 5.9 & 6.2 & 2.8 & 1.0 & 4.2 & 54.0 & 40.0 & 22.0 & 38.1 & 11.3 & 6.8 & 3.0 & 8.0 \\
Grounded-VideoLLM & 12.0 & 8.5 & 5.6 & 9.6 & 20.2 & 12.8 & 7.3 & 15.8 & 54.2 & 36.4 & 19.7 & 36.8 & 46.2 & 30.3 & 19.0 & 36.1 \\
TimeSuite & 1.9 & 0.3 & 0.1 & 2.6 & 17.2 & 8.7 & 4.1 & 11.8 & 69.9 & 48.7 & 24.0 & 45.5 & 31.0 & 16.9 & 9.8 & 23.6 \\
NumPro & 5.2 & 2.8 & 0.8 & 4.8 & 10.4 & 4.7 & 2.6 & 8.1 & 63.8 & 42.0 & 20.6 & 41.4 & 55.6 & 37.5 & 20.6 & 38.8 \\
ED-VTG & -- & -- & -- & -- & -- & -- & -- & -- & 59.5 & 39.3 & 19.8 & 40.2 & 52.1 & 33.1 & 16.0 & 35.2 \\
Time-R1 & 5.7 & 1.9 & 0.8 & 5.5 & 28.4 & 15.4 & 7.8 & 20.7 & \textbf{78.3} & \textbf{61.2} & \textbf{34.9} & \textbf{53.7} & \textbf{58.1} & \underline{39.0} & \textbf{21.4} & \underline{40.5} \\
\midrule
\multicolumn{17}{l}{\textit{Omni-modal Models}} \\
VideoLLaMA2 & 5.1 & 1.6 & 0.9 & 3.3 & 3.3 & 1.5 & 0.0 & 2.4 & 19.4 & 11.5 & 5.0 & 12.9 & 13.0 & 7.1 & 2.8 & 10.2 \\
OLA & 14.9 & 9.9 & 5.3 & 10.1 & 9.6 & 4.2 & 1.9 & 7.5 & 32.2 & 10.9 & 3.2 & 20.0 & 27.9 & 15.6 & 7.0 & 20.6 \\
Qwen2.5-Omni (3B) & 11.5 & 7.0 & 2.9 & 8.4 & 8.6 & 4.6 & 2.0 & 6.1 & 47.4 & 30.9 & 14.4 & 32.2 & 27.5 & 15.7 & 8.5 & 19.9 \\
Qwen2.5-Omni (7B) & 8.4 & 3.9 & 1.2 & 6.0 & 17.9 & 9.2 & 3.8 & 12.3 & 65.5 & 45.9 & 23.3 & 43.7 & 41.5 & 26.0 & 14.6 & 29.7 \\
Qwen3-Omni & 18.2 & 12.3 & 5.2 & 14.4 & 14.4 & 7.7 & 3.4 & 10.4 & 35.4 & 23.4 & 10.5 & 23.3 & 37.6 & 23.9 & 12.6 & 26.9 \\
LongVALE & 4.8 & 2.4 & 1.1 & 4.5 & 15.7 & 8.6 & 3.9 & 11.0 & 47.9 & 19.8 & 9.7 & 32.6 & 46.3 & 28.5 & 14.2 & 31.6 \\
ChronusOmni & 42.3 & 33.7 & 21.9 & 31.2 & \underline{49.7} & 32.5 & 17.6 & \underline{34.5} & 72.2 & 47.0 & 20.8 & 46.0 & 56.0 & 35.5 & 20.1 & 39.4 \\
Gemini 3.1 Pro$^\star$ & 55.0 & 50.5 & 44.7 & 47.0 & 66.2 & 49.0 & 31.0 & 46.6 & 74.6 & 44.4 & 15.6 & 44.2 & 64.1 & 46.6 & 28.7 & 46.9 \\

\midrule
OmniGround(3B)$^\dagger$ & 67.3 & 61.2 & 48.1 & 55.3 & 40.2 & 27.3 & 15.1 & 27.7 & 50.2 & 20.5 & 8.4 & 32.6 & 31.0 & 19.2 & 11.2 & 23.5 \\
OmniGround(7B)$^\dagger$ & \underline{70.3} & \underline{67.6} & \underline{55.7} & \underline{60.4} & 46.2 & \underline{34.5} & \underline{20.2} & 32.7 & 67.2 & 38.5 & 19.7 & 44.6 & 50.4 & 36.5 & 19.2 & 38.3 \\
\textbf{OmniGround(3B)} & 68.3 & 62.2 & 50.5 & 56.9 & 41.4 & 28.6 & 16.1 & 28.6 & 60.5 & 31.2 & 15.1 & 38.8 & 31.7 & 19.6 & 11.1 & 23.7 \\
\textbf{OmniGround(7B)} & \textbf{75.2} & \textbf{72.1} & \textbf{64.9} & \textbf{65.9} & \textbf{51.9} & \textbf{39.5} & \textbf{25.1} & \textbf{37.5} & \underline{74.5} & \underline{52.6} & \underline{27.7} & \underline{49.3} & \underline{56.4} & \textbf{39.1} & \underline{21.1} & \textbf{41.2} \\
\bottomrule
\end{tabular}
\caption{Comparison with state-of-the-art methods on temporal video grounding benchmarks. OmniGround sets a new state of the art on the omni-modal temporal grounding; moreover, despite seeing \emph{no} vision-only grounding data, it still essentially matches specialized methods on the vision-only Charades-STA and ActivityNet-Captions, reflecting strong generalization. \textbf{Bold}: best; \underline{underline}: second best. $^\star$Closed-source, for reference only. $^\dagger$Trained on the same data but with standard SFT (cross-entropy).}
\label{tab:main}
\end{table*}
\textbf{Vision-only temporal grounding.}
The benefit of TACG is not limited to the audio-visual setting. OmniGround never sees any vision-only grounding data during training, and Charades-STA and ActivityNet-Captions are outside the audio-visual grounding domains used for our data construction. Nevertheless, OmniGround remains highly competitive on these benchmarks, essentially matching specialized vision-only methods, trailing only the vision-only specialist Time-R1, and leading all other omni-modal models. This indicates that aligning predictions with internal attention strengthens temporal grounding in general, rather than only fitting the omni-modal training distribution.

\begin{figure}[t]
\centering
\includegraphics[width=0.98\linewidth]{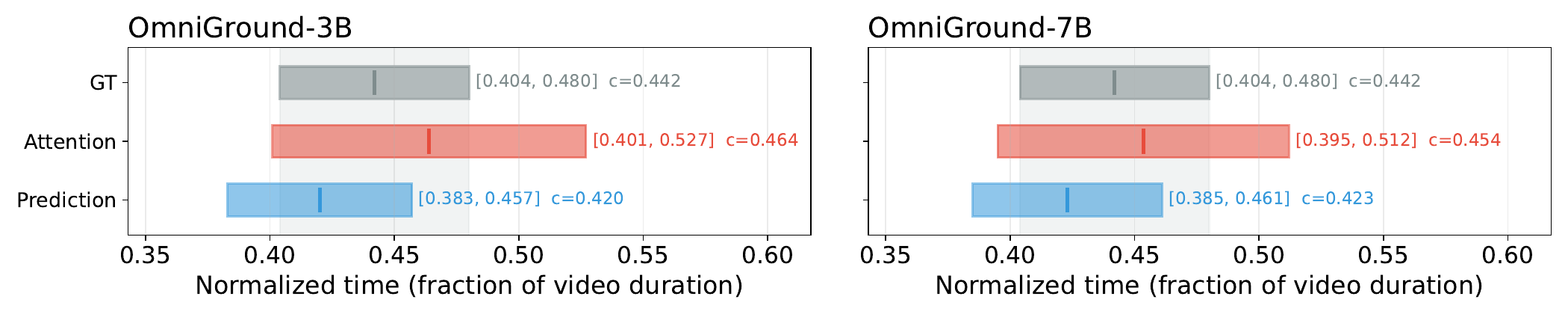}
\caption{\textbf{TACG closes the temporal generation gap.} Average ground-truth, attention, and generated intervals for OmniGround. The persistent displacement observed in the preliminary analysis has been largely absorbed.}
\label{fig:intervals_omniground}
\end{figure}

\begin{table}[t]
\centering
\setlength{\tabcolsep}{8pt}
\small
\begin{tabular}{l|cc}
\toprule
Method & WorldSense & Daily-Omni \\
& Acc & Acc \\
\midrule
VideoLLaMA2 & 27.8 & 35.8 \\
OLA & 34.0 & 54.3 \\
Qwen2.5-Omni (3B) & 46.0 & 62.3 \\
Qwen2.5-Omni (7B) & 46.1 & 62.7 \\
Qwen3-Omni & \textbf{52.0} & 68.8 \\
LongVALE & 9.9 & 24.4 \\
ChronusOmni & 49.5 & \underline{70.2} \\
Gemini 3.1 Pro$^\star$ & 63.1 & 81.0 \\

\midrule
OmniGround (3B)$^\dagger$ & 48.9 & 65.4 \\
OmniGround (7B)$^\dagger$ & 48.3 & 68.7 \\
\textbf{OmniGround (3B)} & 48.6 & 68.1 \\
\textbf{OmniGround (7B)} & \underline{50.2} & \textbf{71.9} \\
\bottomrule
\end{tabular}
\caption{Evaluation on general audio-visual understanding benchmarks. $^\star$Closed-source, for reference only. $^\dagger$Trained on the same data but with standard SFT.}
\label{tab:general}
\end{table}

\textbf{TACG versus standard SFT.}
At full scale, we train otherwise identical 3B and 7B models with standard cross-entropy on the same 580k-sample dataset (OmniGround$^\dagger$); TACG improves grounding on nearly every benchmark at both scales. These full-scale margins are smaller than the clean gain measured in the controlled ablation (Table~\ref{tab:edge_ablation}), and by design: the TACG objective activates only on the forward-TVG samples---about one-fifth of the 580k-sample mixture---so its effect is diluted at the full-model level. Even under this dilution, the improvement is consistent across the evaluated benchmarks, and on the never-seen vision-only Charades-STA it is pronounced at both scales (+6.2 mIoU at 3B), showing that aligning predictions with the internal attention instills a \emph{transferable} localization skill rather than overfitting a particular distribution. Figure~\ref{fig:intervals_omniground} makes this alignment visible on AVID: for both OmniGround scales, the predicted intervals become better aligned with attention and move closer to the ground-truth intervals, explaining the stronger localization performance and confirming that TACG closes the temporal generation gap it was designed to target.

\subsection{General Understanding Performance}

Table~\ref{tab:general} shows that OmniGround preserves strong general audio-visual understanding while improving temporal grounding. The 7B model remains competitive on WorldSense and achieves the best open-model result on Daily-Omni, indicating that grounding-oriented training does not come at the cost of general comprehension. Relative to the same-scale cross-entropy baselines, TACG keeps WorldSense stable and further improves Daily-Omni, suggesting that sharper temporal alignment can also benefit tasks involving event order and audio-visual correspondence. By contrast, the grounding-specialized LongVALE performs poorly on these general benchmarks, highlighting the advantage of OmniGround's balanced training.

%% file: sections/ablation.tex
\section{Ablation Study}
\label{sec:ablation}

We conduct ablation experiments to analyze the key design choices in TACG. All ablations are trained on the same set of 10k TVG samples, with an SFT model trained on identical data as the baseline, so the comparison isolates the gains brought by TACG rather than data differences.

\subsection{Component Ablation}
We evaluate the contribution of each consistency edge in Eq.~\ref{eq:total_loss} ($\mathcal{L}_{\text{tg}}$, $\mathcal{L}_{\text{pg}}$, and $\mathcal{L}_{\text{tp}}$). Each configuration is reported under its individually optimized weight, so the comparison reflects each edge's best achievable effect.

\begin{table}[t]
\centering
\setlength{\tabcolsep}{4pt}
\small
\begin{tabular}{l|cc|cc|r}
\toprule
\multirow{2}{*}{Configuration} & \multicolumn{2}{c|}{AVID} & \multicolumn{2}{c|}{LongVALE} & \multirow{2}{*}{$\Delta$\%} \\
& R@0.5 & mIoU & R@0.5 & mIoU & \\
\midrule
Baseline (CE only) & 0.3502 & 0.3153 & 0.1408 & 0.1604 & 0.00 \\

$\mathcal{L}_{\text{tg}}$ only & 0.3543 & 0.3223 & \textbf{0.1563} & \underline{0.1722} & \underline{+4.79} \\
$\mathcal{L}_{\text{pg}}$ only & 0.3649 & 0.3282 & 0.1479 & 0.1645 & +3.32 \\
$\mathcal{L}_{\text{tp}}$ only & \underline{0.3671} & \underline{0.3315} & 0.1514 & 0.1663 & +4.41 \\[2pt]

$\mathcal{L}_{\text{tg}} + \mathcal{L}_{\text{pg}}$ & 0.3671 & 0.3301 & 0.1508 & 0.1672 & +4.47 \\[4pt]

\textbf{Full TACG} & \textbf{0.3814} & \textbf{0.3408} & \underline{0.1546} & \textbf{0.1759} & \textbf{+8.87} \\
\bottomrule
\end{tabular}
\caption{Component ablation of the TACG consistency loss. $\Delta$\% is the average relative improvement over the SFT baseline across both benchmarks. \textbf{Bold}: best; \underline{underline}: second best. Full results are provided in the appendix.}
\label{tab:edge_ablation}
\end{table}

As shown in Table~\ref{tab:edge_ablation}, each single edge improves over the SFT baseline under its own optimal weight (+3.32\% to +4.79\%), indicating that all three consistency signals are useful on their own.
Stacking the two ground-truth-anchored edges ($\mathcal{L}_{\text{tg}} + \mathcal{L}_{\text{pg}}$) yields only +4.47\%---essentially no gain over either edge alone. In sharp contrast, completing the triangle with the attention-to-prediction edge $\mathcal{L}_{\text{tp}}$ lifts the gain to +8.87\%, nearly doubling the best two-edge result. This identifies attention--prediction consistency as the core component of TACG: the largest gain appears precisely when the objective addresses the temporal generation gap diagnosed in our preliminary analysis.

\subsection{Discretization of the Predicted Distribution}

Constructing the predicted distribution requires mapping the per-position digit logits to a digit value $\tilde{d}_i$, which is then assembled into a timestamp by place value. Two conventional choices both fail for timestamp regression: \emph{Argmax} keeps only the most probable digit, so it cannot distinguish a nearly-correct distribution from a confidently-wrong one and its selected digit jumps discontinuously between steps, yielding unstable gradients; \emph{Soft} uses the continuous expectation for smooth gradients, but the expectation of a discrete distribution rarely lands on an integer, leaving a loss floor that, amplified by place value, biases the gradient at high-order positions.

We therefore adopt \emph{Round}: we first take the soft expectation and then round it to the nearest digit. This combines smooth gradients with discrete semantics, avoiding both Argmax's instability and Soft's vanishing-gradient floor.
\begin{figure}[t]
\centering
\includegraphics[width=0.49\linewidth]{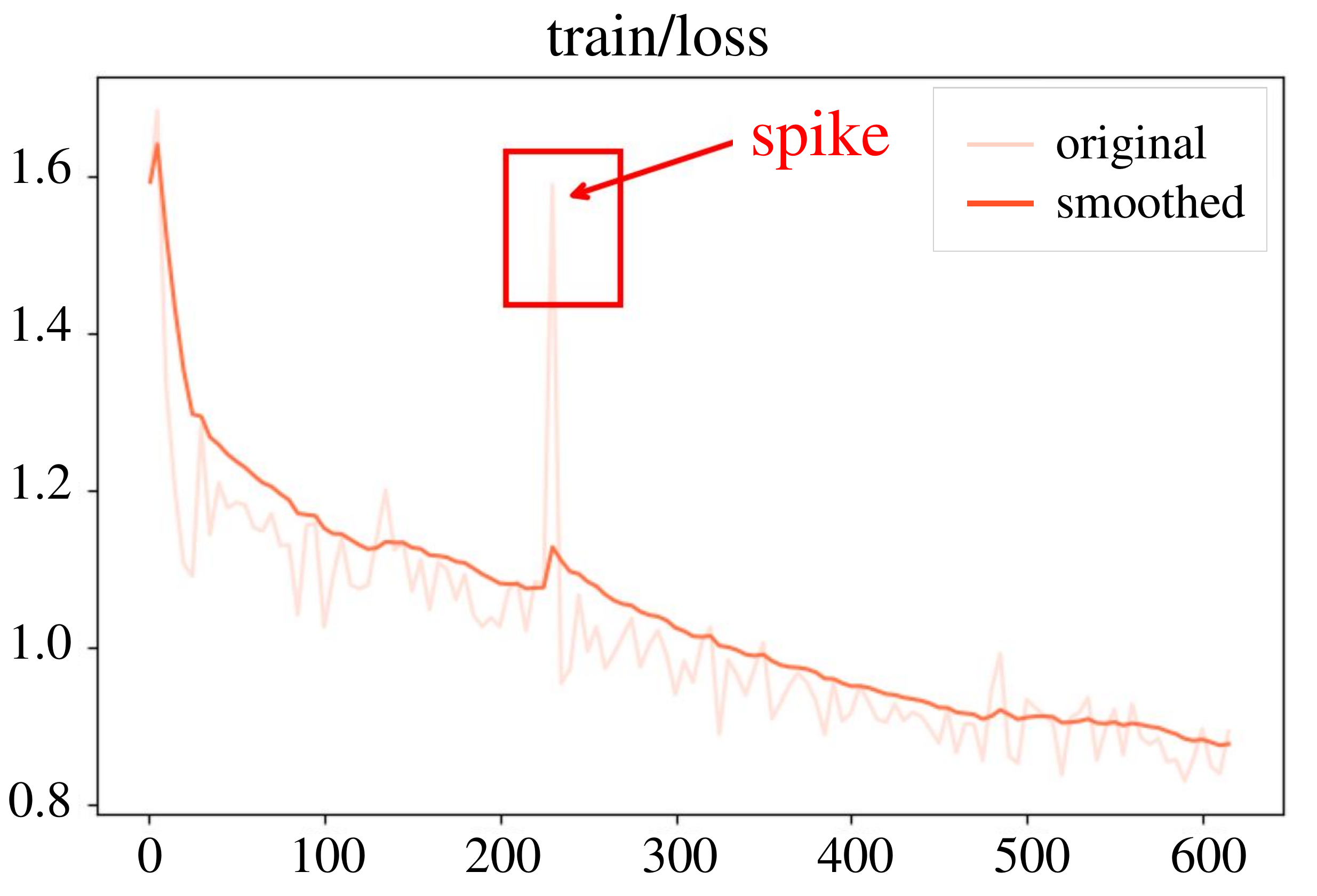}\hfill
\includegraphics[width=0.49\linewidth]{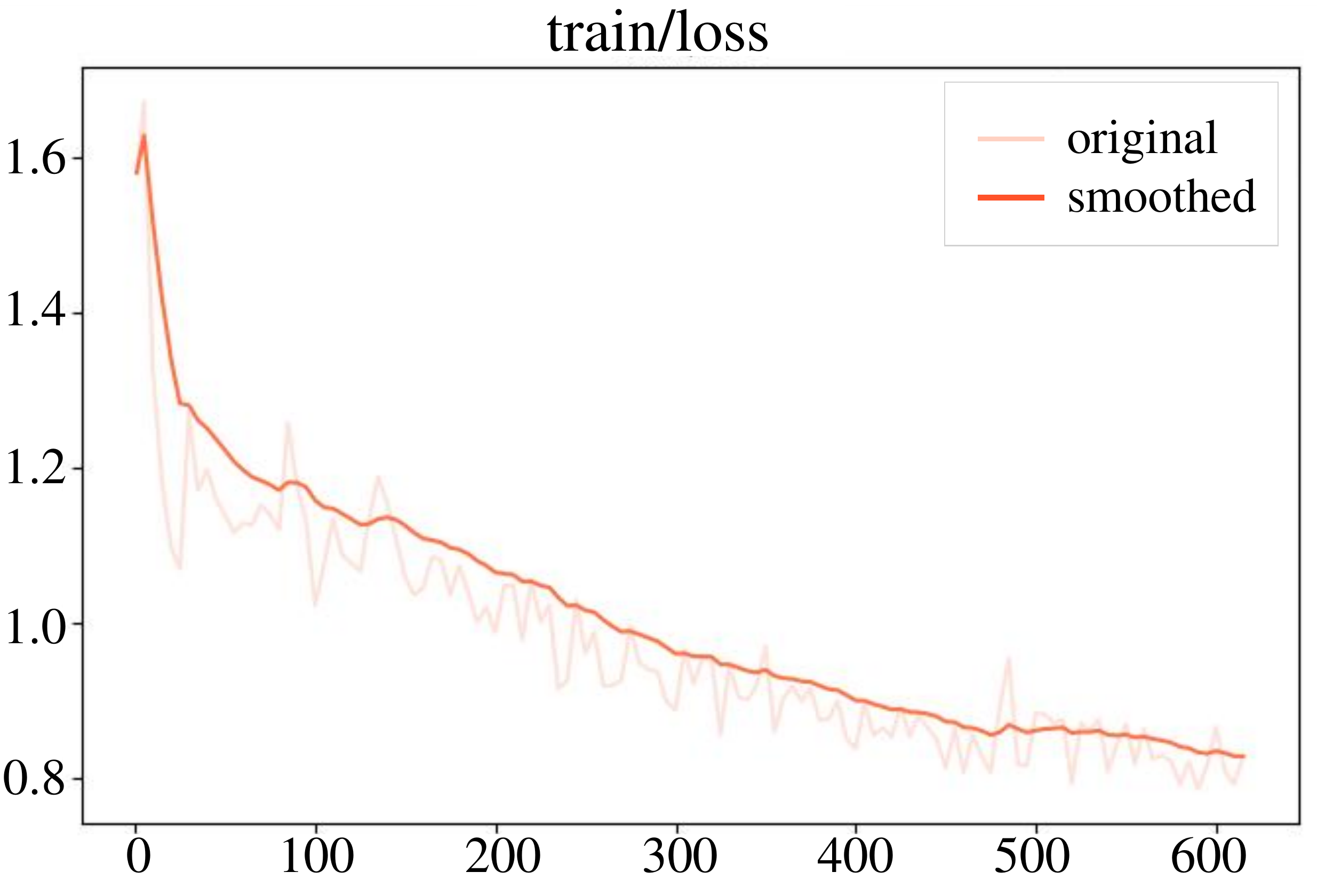}
{\footnotesize (a) Argmax loss \hspace{5em} (b) Round loss}
\caption{Training dynamics for timestamp discretization. Argmax spikes mid-training from mode collapse, whereas Round decreases smoothly and monotonically.}
\label{fig:discretization_training}
\end{figure}
The metrics confirm these failure modes: both Argmax and Soft fall below the CE baseline at their best settings, while Round is the only choice that yields a stable improvement (Appendix Figure~\ref{fig:discretization_bar}). The training dynamics tell the same story (Figure~\ref{fig:discretization_training}): Argmax shows a sharp mid-training spike from mode collapse, whereas Round decreases smoothly and monotonically. The per-$\lambda$ sensitivity curves are reported in the appendix.

%% file: sections/conclusion.tex
\section{Conclusion}
\label{sec:conclusion}
We identified the \textit{temporal generation gap} in omni-modal language models, where internal attention localizes the target moment more accurately than the generated timestamps, and proposed Temporal Attention-Consistency Grounding (TACG), which couples attention, prediction, and ground truth into a triangle of consistency constraints that strengthens temporal localization with no added parameters, special tokens, or inference overhead. We further constructed a 580k-sample training dataset covering both grounding and general omni-modal understanding. Built on TACG and this dataset, OmniGround reaches state-of-the-art omni-modal temporal grounding while preserving strong general comprehension. More broadly, our results show that a model's internal attention is a valuable and underexploited supervision signal for temporal precision; we will release the dataset, model, and training framework to support future work.

%% file: sections/appendix.tex
\clearpage
\appendix
\begin{center}
{\Large\bfseries Technical Appendix}
\end{center}
\vspace{0.5em}

\noindent
This appendix provides additional details omitted from the main text due to space constraints. It is organized into three parts: (i) supporting material for the preliminary analysis of the temporal generation gap (divergence definitions and per-layer attention analysis); (ii) model architecture and training details; and (iii) additional ablations and full experimental results.


\section{Divergence Definitions}
\label{app:divergences}

For two distributions $p, q$ over the $T$ time steps, the Jensen--Shannon divergence and the Wasserstein-1 distance used in the consistency loss (Eq.~\ref{eq:total_loss}) are
\begin{equation}
\begin{aligned}
\mathrm{JS}(p, q) &= \tfrac{1}{2}\mathrm{KL}(p \,\|\, m) + \tfrac{1}{2}\mathrm{KL}(q \,\|\, m), \quad m = \tfrac{1}{2}(p+q), \\
\mathcal{W}_1(p, q) &= \frac{1}{T}\sum_{t=1}^{T}\Big\lvert \textstyle\sum_{\tau \le t} (p_\tau - q_\tau) \Big\rvert,
\end{aligned}
\end{equation}
where $\mathrm{KL}$ is the Kullback--Leibler divergence and the second expression is the $\ell_1$ distance between the cumulative distributions, the closed form of the Wasserstein-1 distance in one dimension.

\section{Per-Layer Attention Analysis}
\label{app:layer_analysis}

\begin{figure*}[t]
\centering
\includegraphics[width=\linewidth]{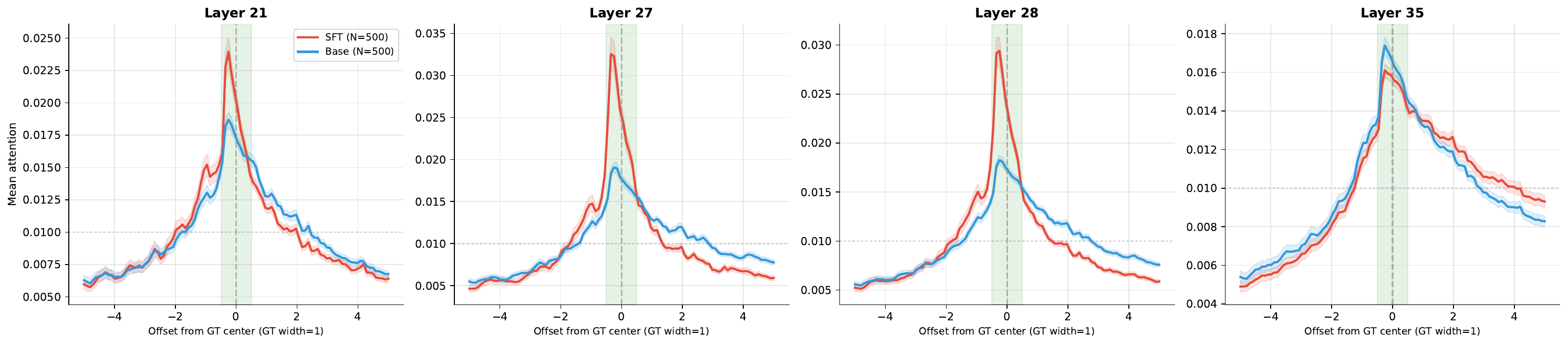}
\caption{GT-aligned attention at representative layers for the SFT and base models (N=500); the green band marks the ground-truth segment.}
\label{fig:three_dist}
\end{figure*}

\begin{figure*}[t]
\centering
\includegraphics[width=\linewidth]{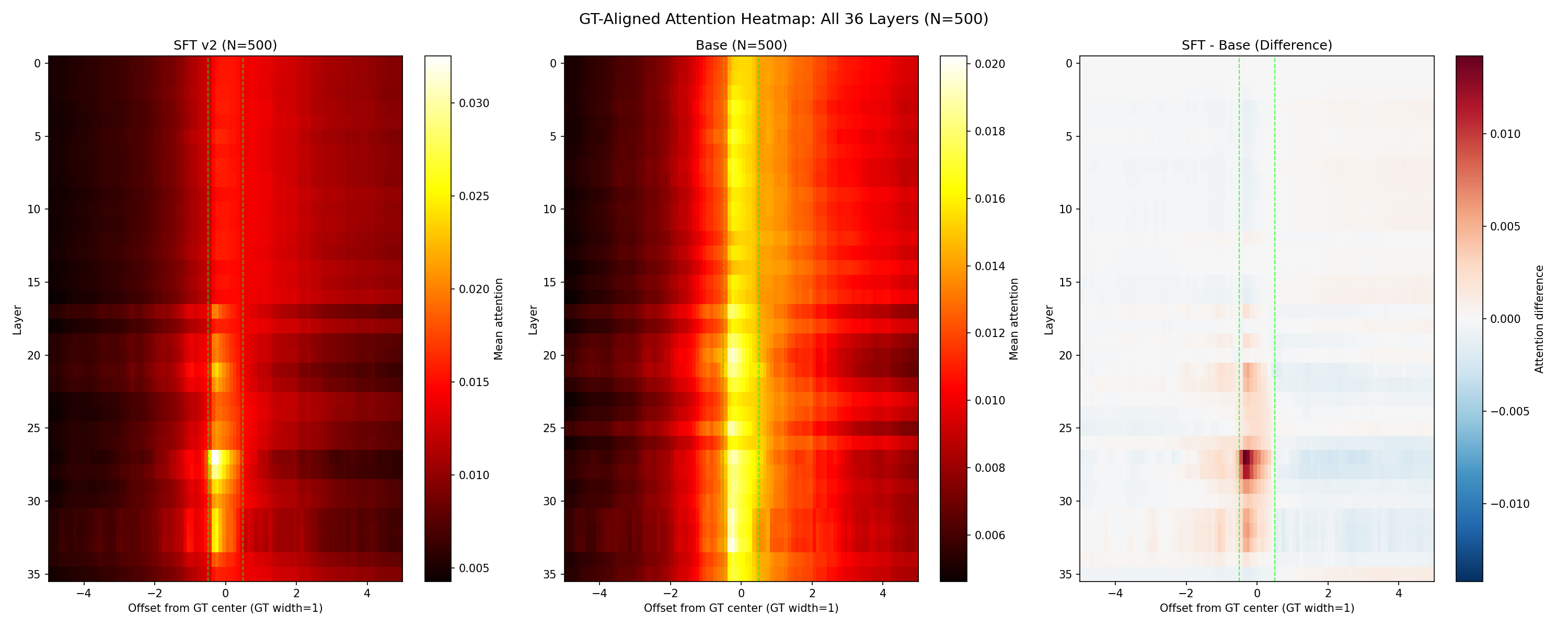}
\caption{GT-aligned temporal attention across all 36 layers ($N{=}500$), for the SFT model (left), the base model (middle), and their difference (right). Each row is a layer; the horizontal axis is the offset from the ground-truth center (GT width normalized to 1; green dashed lines mark the segment). Attention concentrates inside the GT band at nearly every layer and is sharpest around layers 26--28, and the SFT$-$base difference peaks inside the band at the same layers---fine-tuning sharpens these mid-to-late layers the most.}
\label{fig:layer_heatmap}
\end{figure*}

We obtain each layer's audio-visual temporal attention distribution from the last prefill token. We take its attention weights normalized over all video and audio tokens, assign both modalities to their corresponding interleaved temporal grids, sum the attention mass within each grid, and average over heads, yielding the distribution $p_{\text{att}}$ over the $T$ media grids. To make distributions comparable across clips of different length and segment position, we re-align every distribution so that the ground-truth segment is centered at the origin and rescaled to unit width, then average within each layer.

For each benchmark, the diagnostic evaluation split consists of 500 examples randomly sampled from the official test split. All preliminary-analysis statistics in Table~\ref{tab:js_gap}, including Attn.\ wins and prediction mIoU, are computed on these fixed benchmark-specific subsets; ties are excluded only from the denominator of Attn.\ wins. The layer-wise visualization and sensitivity analysis use the same sampling protocol.

\begin{table}[t]
\centering
\setlength{\tabcolsep}{4pt}
\small
\begin{tabular}{ll|ccc}
\toprule
Variant & Group & Attn.\ mIoU & Pred.\ mIoU & $\Delta$IoU \\
\midrule
\multirow{4}{*}{CE Base}
& Attn.\ wins & 40.0 & 5.9 & +34.1 \\
& Pred.\ wins & 37.9 & 68.5 & $-$30.6 \\
& Ties & 0.0 & 0.0 & 0.0 \\
& \textbf{Overall} & \textbf{29.8} & \textbf{31.1} & \textbf{$-$1.3} \\
\midrule
\multirow{4}{*}{Full TACG}
& Attn.\ wins & 50.0 & 8.0 & +42.0 \\
& Pred.\ wins & 41.1 & 74.0 & $-$32.9 \\
& Ties & 0.0 & 0.0 & 0.0 \\
& \textbf{Overall} & \textbf{34.6} & \textbf{34.1} & \textbf{+0.5} \\
\bottomrule
\end{tabular}
\caption{Attention--prediction alignment of the CE baseline and Full TACG on AVID. All values are reported on the percentage scale. Full TACG improves both localization signals and reduces the aggregate gap between attention and prediction.}
\label{tab:ce_tacg_alignment}
\end{table}

\noindent\textbf{Lead rate versus aggregate mIoU.}
These two statistics characterize different aspects of localization quality. Lead rate measures how frequently attention is closer to the ground truth than the generated prediction at the instance level, whereas aggregate mIoU is also sensitive to the magnitude of a smaller number of large wins. Consequently, attention can lead on more non-tied samples while having a similar or slightly lower aggregate mIoU, as observed for the CE baseline. Our claim is therefore not that attention is uniformly superior on every sample, but that it provides complementary temporal evidence on a substantial fraction of instances. TACG exploits this evidence while anchoring both attention and prediction to the ground truth.

We visualize all 36 transformer layers to examine how the audio-visual temporal-localization signal evolves with model depth. As Figure~\ref{fig:three_dist} shows, layers 27--28 exhibit both sharp attention concentration inside the ground-truth segment and substantial sharpening from supervised fine-tuning, whereas earlier layers (e.g.\ layer~21) are less discriminative and the final layer~35 is diffuse. Figure~\ref{fig:layer_heatmap} corroborates this across all 36 layers at once: in both the base and SFT models the attention mass concentrates inside the ground-truth band at essentially every layer, showing that the localization signal is broad rather than confined to a single layer. Throughout our experiments, we use layer~27 of Qwen2.5-Omni-3B (36 layers), layer~22 of Qwen2.5-Omni-7B, layer~32 of the Qwen3-Omni thinker (48 layers), and layer~20 of OLA (28 layers). For TACG training, the fixed attention layer is layer~27 for the 3B model and layer~22 for the 7B model, as reported in Table~\ref{tab:training_hparams}.
The absolute index differs across models only because their depths do, not the underlying phenomenon.

\begin{figure}[t]
\centering
\includegraphics[width=0.49\linewidth]{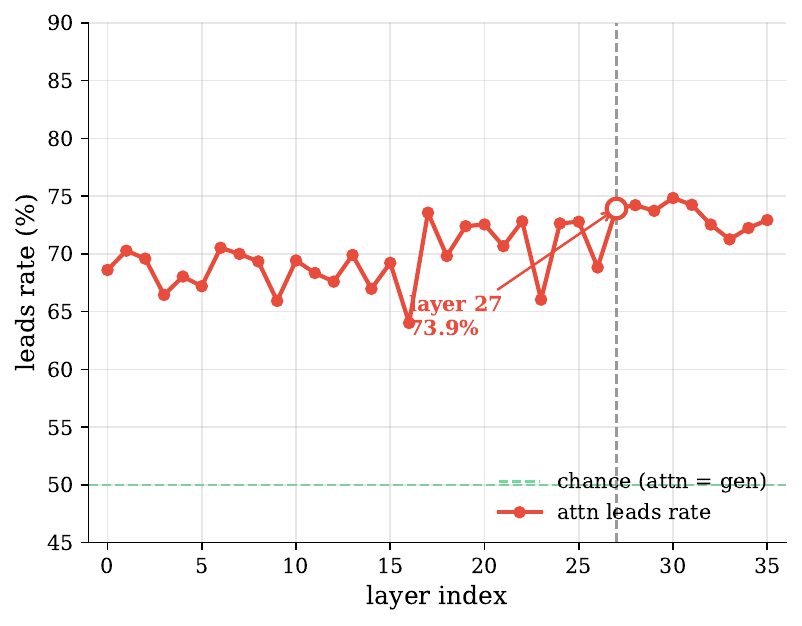}
\includegraphics[width=0.49\linewidth]{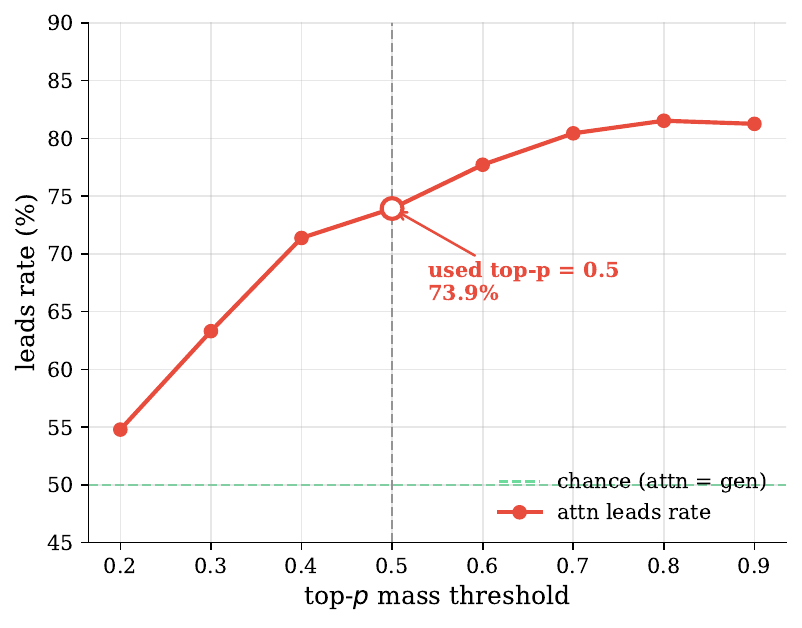}
\caption{Sensitivity of the temporal generation gap to attention extraction choices on Qwen2.5-Omni-3B ($N{=}500$). Left: layer sweep. Right: cumulative-mass cutoff sweep for extracting $I_{\text{att}}$.}
\label{fig:attn_extraction_sensitivity}
\end{figure}

We further test whether the temporal generation gap depends on the attention-extraction hyperparameters. Figure~\ref{fig:attn_extraction_sensitivity} varies the attention layer and the cumulative-mass cutoff used to convert attention into an interval. Across both sweeps, attention still overlaps the ground truth better than the generated interval on most non-tie samples, showing that the gap is not an artifact of a single layer or the default $50\%$ cutoff.

\begin{table}[t]
\centering
\setlength{\tabcolsep}{3pt}
\small
\begin{tabular}{l|l}
\toprule
\textbf{Component} & \textbf{Specification} \\
\midrule
\multicolumn{2}{l}{\textit{Vision Encoder}} \\
\quad Architecture & Vision Transformer (ViT) \\
\quad Hidden size & 1280 \\
\quad Patch size & 14 $\times$ 14 \\
\quad Temporal patch size & 2 \\
\quad Spatial merge size & 2 \\
\quad Input channels & 3 \\
\midrule
\multicolumn{2}{l}{\textit{Audio Encoder}} \\
\quad Architecture & Whisper-style Transformer \\
\quad Hidden size & 1280 \\
\quad Output dimension & 2048 \\
\quad Number of layers & 32 \\
\quad Attention heads & 20 \\
\quad Mel bins & 128 \\
\midrule
\multicolumn{2}{l}{\textit{Language Model}} \\
\quad Architecture & Transformer (autoregressive) \\
\quad Hidden size & 2048 \\
\quad Number of layers & 36 \\
\quad Attention heads & 16 \\
\quad Key-value heads & 2 (grouped-query attention) \\
\quad Intermediate size & 11008 \\
\quad Vocabulary size & 151936 \\
\quad Position encoding & 3D M-RoPE (T, H, W) \\
\midrule
\multicolumn{2}{l}{\textit{Chunk Configuration}} \\
\quad Seconds per chunk & 2 \\
\quad Max temporal grids & 30 \\
\quad Long video threshold & 60 seconds \\
\bottomrule
\end{tabular}
\caption{Detailed architecture specifications of OmniGround.}
\label{tab:arch_detail}
\end{table}
\section{Architecture Details}
\label{app:architecture}

The vision encoder processes input frames at native resolution. Each frame is divided into $14 \times 14$ pixel patches. A temporal patch size of 2 groups two consecutive frames into a single temporal unit, reducing the temporal token count by half. After ViT encoding, a spatial merge with factor 2 further compresses spatial tokens by merging $2 \times 2$ adjacent spatial patches, resulting in a $4\times$ reduction in spatial token count.

\textbf{Audio Encoder.}
Raw audio waveforms are resampled to 16 kHz and converted to 128-bin mel-spectrograms. The Whisper-style encoder uses 32 transformer layers with 20 attention heads and a hidden dimension of 1280. A linear projection maps the output to 2048 dimensions, matching the language model hidden size. Audio longer than 30 seconds is segmented and encoded in parallel, with outputs concatenated along the sequence dimension.

\textbf{Chunk-Interleaved Input Construction.}
For a video of duration $D$ seconds:
\begin{itemize}
\item If $D \leq 60$s: the video is encoded with $\lfloor D / 2 \rfloor$ temporal grids, each covering exactly 2 seconds. Audio tokens are split into chunks of fixed size matching the 2-second interval. Video and audio chunks are interleaved in chronological order.
\item If $D > 60$s: video frames are uniformly sampled and encoded into at most 30 temporal grids. Each grid covers $D / 30$ seconds. Audio tokens are uniformly redistributed into 30 chunks of equal size, ensuring each audio chunk aligns temporally with its corresponding video grid.
\end{itemize}

The interleaved sequence is wrapped with special audio start and end tokens. Within each chunk, visual tokens appear first, followed by audio tokens. The 3D M-RoPE assigns temporal, height, and width position indices to visual tokens, while audio tokens receive sequential position indices synchronized with the temporal position of their chunk.

\section{Training Data Details}
\label{app:data}

\textbf{Quality Filtering.}
Before reformatting, we remove low-resolution videos and overly short clips: grounding videos shorter than 30\,s and general-understanding clips shorter than 2\,s are discarded. General-understanding data is then sampled from four sources so that no single audio type dominates: OLA-reannotated LLaVA-Video for paired speech and visual reasoning, original LLaVA-Video for general video proficiency with audio, AVQA for scene sound, and FortisAVQA for music.
\begin{table}[htbp]
\centering
\setlength{\tabcolsep}{3pt}
\small
\begin{tabular}{l|l}
\toprule
\textbf{Hyperparameter} & \textbf{Value} \\
\midrule
Base model & Qwen2.5-Omni-3B / 7B \\
Optimizer & AdamW \\
Learning rate & 2.5\text{e-}5 \\
LR scheduler & Cosine \\
Warmup ratio & 0.05 \\
Batch size (per device) & 1 \\
Gradient accumulation & 2 \\
Number of epochs & 1 \\
Max sequence length & 16{,}384 \\
DeepSpeed & ZeRO-2 \\
Gradient checkpointing & Enabled \\
Attention implementation & FlashAttention \\
Precision & BF16 \\
Frozen modules & ViT, audio aligner \\
\midrule
\multicolumn{2}{l}{\textit{TACG Configuration}} \\
\quad $\lambda_{\text{tp}}$ (attn--pred, $\mathcal{W}_1$) & 0.0500 \\
\quad $\lambda_{\text{tg}}$ (attn--GT, JS) & 0.0125 \\
\quad $\lambda_{\text{pg}}$ (pred--GT, $\mathcal{W}_1$) & 0.1500 \\
\quad Attention layer $l$ & 27 (3B) / 22 (7B) \\
\quad Gaussian $\sigma$ & $0.15 \times \text{segment width}$ \\
\bottomrule
\end{tabular}
\caption{Training hyperparameters and TACG configuration.}
\label{tab:training_hparams}
\end{table}

\textbf{QA Construction.}
AVID and LongVALE provide source videos with dense caption-style temporal annotations rather than the final QA pairs used for our training. We use only their training-split source videos and temporal captions, then reconstruct new plain-text QA instances in three complementary grounding formats. We preserve the official benchmark separation: videos, temporal annotations, and captions from the held-out evaluation splits are excluded from all construction, filtering, and training stages. \emph{Forward TVG} maps an event description to its interval---\texttt{``Locate the event where \{description\}.''} $\rightarrow$ \texttt{``[start\_sec, end\_sec]''}; \emph{Reverse Grounding} inverts this, mapping a queried time interval to the event that occurs there; and \emph{Dense Video Captioning} asks the model to enumerate all timed events in a clip---\texttt{``[start, end]: \{event\}''} lines in chronological order. This trains a single model to move in both directions between language and time rather than only emitting timestamps.

\textbf{Audio-Visual Relevance Filtering and Mixing.}
After reformatting we apply an audio-visual relevance filter at the individual caption/query level, retaining targets that describe audio-visual events and removing text targets that are audio-only, visual-only, or ambiguous. For TVG and Reverse Grounding, the target is the queried event caption; for DVC, each timed event caption in the same QA instance is judged independently. The filter is text-only: Gemini~3.1~Pro receives only the event caption or QA text as input, labels the target as audio-visual, audio-only, visual-only, or ambiguous, and retains only audio-visual cases. Rejection rates are therefore computed over individual caption/query targets rather than whole videos. The filter is applied only to training-split samples and never sees held-out evaluation videos, annotations, or captions. A final balanced-mixing step combines the two streams into a 580,676-sample dataset at a grounding-to-general ratio of roughly $1{:}2$, which we found preserves general audio-visual understanding while providing enough grounding supervision for the TACG objective to take effect.

\begin{table}[h]
\centering
\setlength{\tabcolsep}{6pt}
\small
\begin{tabular}{l|ccc}
\toprule
\textbf{Rejected type} & \textbf{Visual-only} & \textbf{Audio-only} & \textbf{Total} \\
\midrule
\textbf{Rate} & $17\%$ & $4\%$ & $21\%$ \\
\bottomrule
\end{tabular}
\caption{Text-only audio-visual relevance filtering statistics. Rejection is computed over individual caption/query targets rather than whole videos or QA instances.}
\label{tab:av_filter_stats}
\end{table}

\begin{table}[h]
\centering
\setlength{\tabcolsep}{2pt}
\footnotesize
\begin{tabular}{l|l|c}
\toprule
\textbf{Task} & \textbf{Source} & \textbf{Samples} \\
\midrule
\multicolumn{3}{l}{\textit{Temporal Grounding Data (from AVID \& LongVALE)}} \\
\quad Event $\rightarrow$ Time (TVG) & AVID & 34,041 \\
\quad Event $\rightarrow$ Time (TVG) & LongVALE & 91,863 \\
\quad Time $\rightarrow$ Event (Reverse) & AVID & 13,526 \\
\quad Time $\rightarrow$ Event (Reverse) & LongVALE & 36,474 \\
\quad Dense Captioning (DVC) & AVID & 7,239 \\
\quad Dense Captioning (DVC) & LongVALE & 7,240 \\
\quad \textit{Subtotal} & & \textit{190,383} \\
\midrule
\multicolumn{3}{l}{\textit{General Understanding Data}} \\
\quad General video QA & LLaVA-Video & 200,000 \\
\quad Speech\,+\,visual QA & LLaVA-Video (OLA) & 120,925 \\
\quad Scene-sound QA & AVQA & 37,441 \\
\quad Music QA & FortisAVQA & 31,927 \\
\quad \textit{Subtotal} & & \textit{390,293} \\
\midrule
\multicolumn{2}{l|}{\textbf{Total}} & \textbf{580,676} \\
\bottomrule
\end{tabular}
\caption{Training data composition of our final model (580,676 samples). Reverse Grounding is produced by inverting a 50{,}000-sample subset of the forward-TVG pool, so it inherits that pool's AVID/LongVALE split. The SFT model probed in our preliminary analysis is trained \emph{only} on the forward-TVG data (125,904 samples), without the Reverse Grounding, DVC, or general-understanding data.}
\label{tab:data_detail}
\end{table}

\section{Training Hyperparameters}
\label{app:training}

We fine-tune Qwen2.5-Omni-3B and Qwen2.5-Omni-7B in a single stage with AdamW and a cosine schedule, freezing the vision encoder and audio aligner so that training adapts the language model and cross-modal fusion while preserving the pretrained perceptual front-ends. Both scales use the same optimization and TACG configurations except for the attention layer specified in Table~\ref{tab:training_hparams}. The TACG objective is applied selectively---all forward-TVG samples, which emit an explicit interval and thus admit a predicted distribution, use the full consistency loss, while all other samples use cross-entropy alone. The three edge weights $(\lambda_{\text{tp}},\lambda_{\text{tg}},\lambda_{\text{pg}})$ are the effective values used by the final models.

\textbf{Computing Infrastructure and Software.}
The full computing environment is listed in Table~\ref{tab:infra}. Main-text results use a fixed random seed (42); multi-seed mean$\pm$std for the key results is reported in the Multi-Seed Stability section (Tables~\ref{tab:multiseed_main} and~\ref{tab:multiseed_ablation}). Code, training data, and model checkpoints will be released publicly upon publication.

\begin{table}[t]
\centering
\setlength{\tabcolsep}{2pt}
\small
\begin{tabular}{l|l}
\toprule
\textbf{Item} & \textbf{Specification} \\
\midrule
\multicolumn{2}{l}{\textit{Hardware}} \\
\quad GPUs & 32 $\times$ NVIDIA A100-80GB \\
\quad Parallelism & DeepSpeed ZeRO-2 \\
\quad NVIDIA driver & 470.161.03 \\
\midrule
\multicolumn{2}{l}{\textit{Software}} \\
\quad Python & 3.10.15 \\
\quad PyTorch & 2.8.0 (CUDA 12.8) \\
\quad \texttt{transformers} & 4.57.0 \\
\quad \texttt{ms-swift} & 4.2.0 \\
\quad \texttt{accelerate} & 1.13.0 \\
\quad \texttt{deepspeed} & 0.17.2 \\
\quad \texttt{peft} & 0.15.0 \\
\quad \texttt{flash-attn} & 2.8.1 \\
\quad \texttt{xformers} & 0.0.32 \\
\quad \texttt{triton} & 3.4.0 \\
\quad \texttt{datasets} & 4.8.4 \\
\quad \texttt{tokenizers} & 0.22.1 \\
\quad \texttt{torchaudio} / \texttt{torchvision} & 2.8.0 / 0.23.0 \\
\quad \texttt{vllm} (inference/eval) & 0.10.2 \\
\bottomrule
\end{tabular}
\caption{Computing infrastructure and software environment.}
\label{tab:infra}
\end{table}

\begin{figure}[htbp]
\centering
\includegraphics[width=0.82\linewidth]{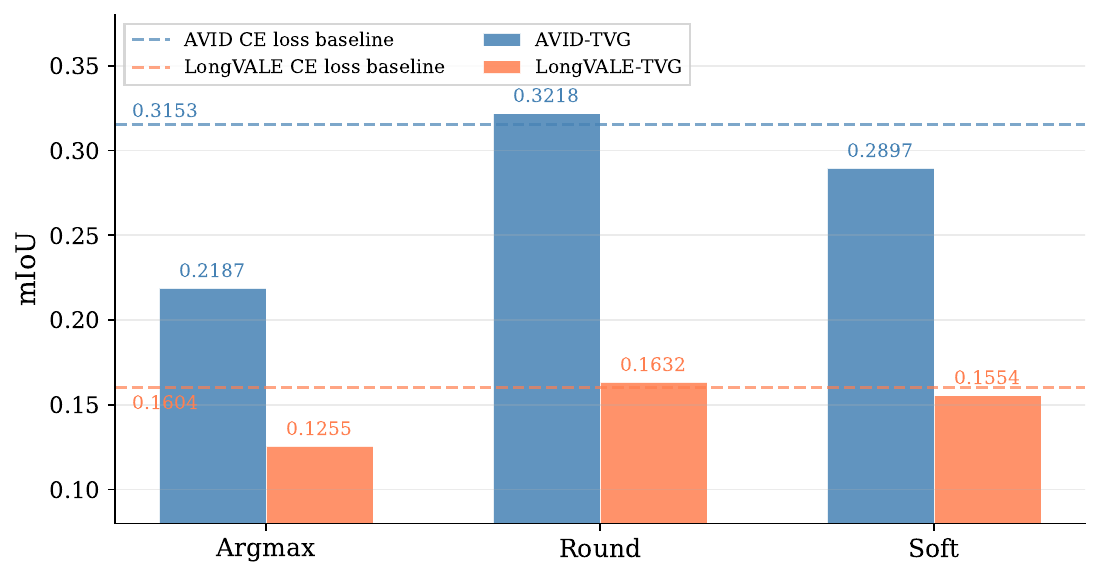}
\caption{Best mIoU per discretization against the cross-entropy baseline. Only Round yields a positive gain over the CE baseline; Argmax and Soft fall below it on both benchmarks.}
\label{fig:discretization_bar}
\end{figure}

\begin{figure}[htbp]
\centering
\includegraphics[width=\linewidth]{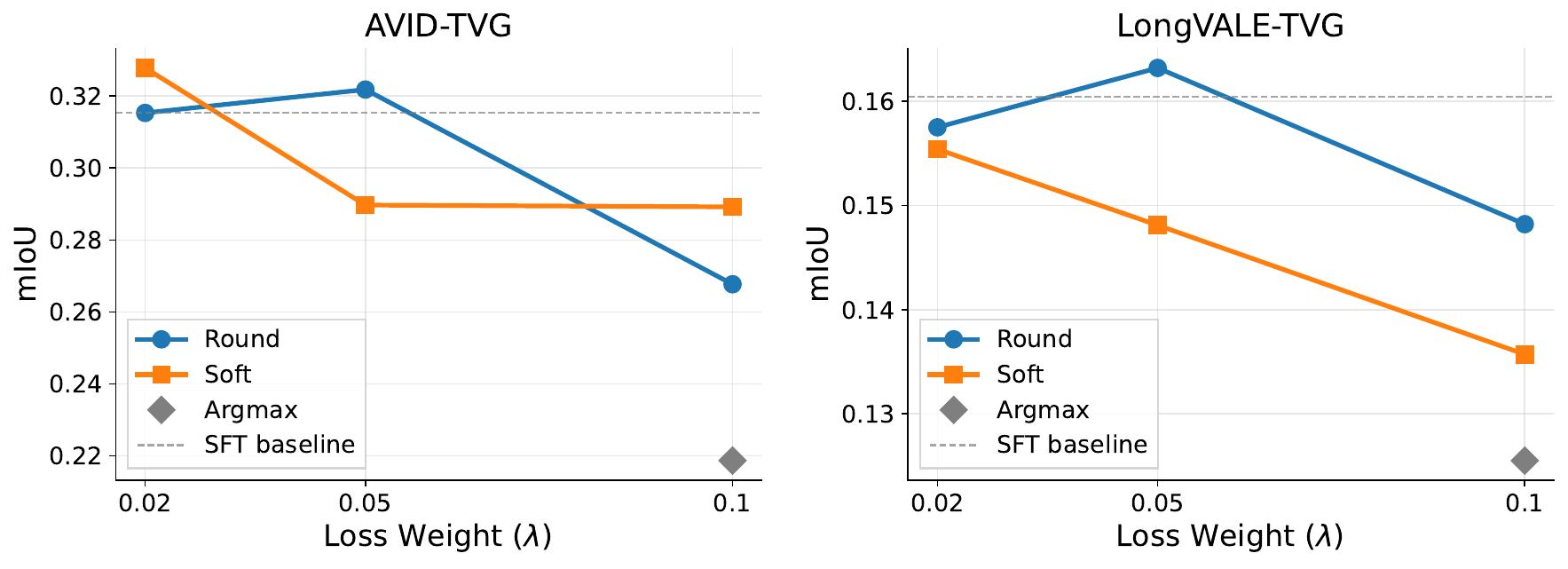}
\caption{mIoU vs.\ loss weight $\lambda$ for each discretization.}
\label{fig:discretization_curve}
\end{figure}

\begin{figure}[htbp]
\centering
\includegraphics[width=\linewidth]{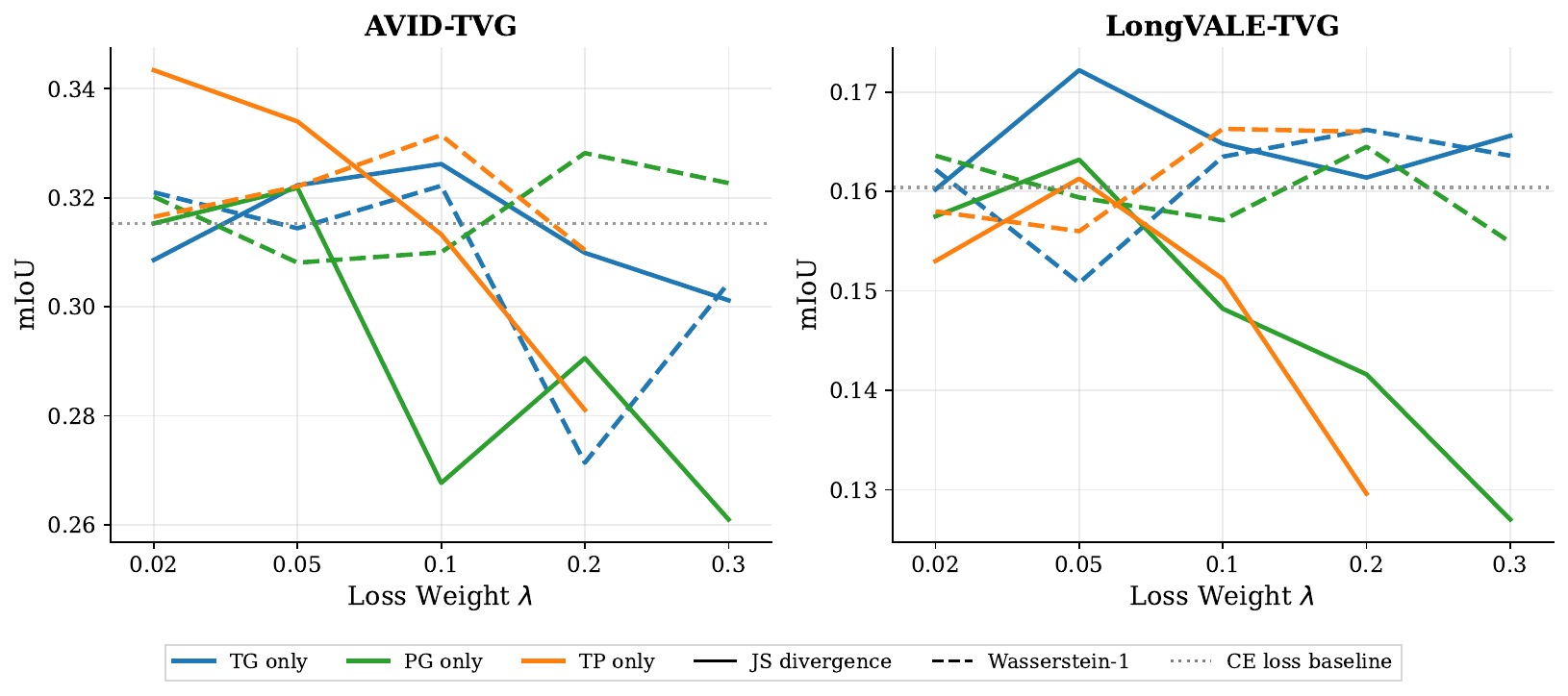}
\caption{Weight ablation for all three edges ($\mathcal{L}_{\text{tg}}$, $\mathcal{L}_{\text{pg}}$, $\mathcal{L}_{\text{tp}}$) on both benchmarks.}
\label{fig:weight_sensitivity}
\end{figure}

\begin{figure}[htbp]
\centering
\includegraphics[width=\linewidth]{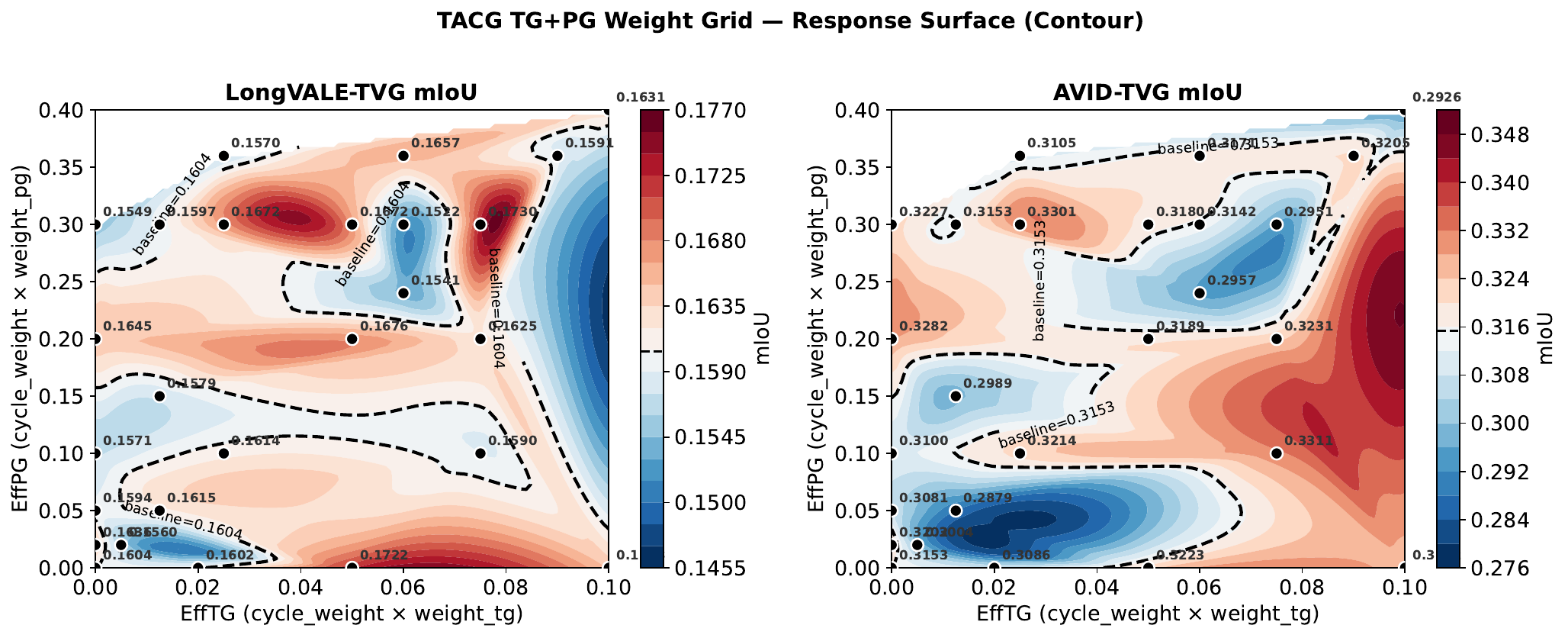}
\caption{Joint mIoU response surface over the $\lambda_{\text{tg}}$--$\lambda_{\text{pg}}$ grid (26 evaluated points; dashed contour = SFT baseline) on LongVALE (left) and AVID (right).}
\label{fig:tgpg_surface}
\end{figure}

\section{Evaluation Protocol and Baseline Reporting}
\label{app:evaluation_protocol}

All grounding results are evaluated under the official protocol of each benchmark. For AVID and LongVALE, we use the released test splits and compute R@0.3, R@0.5, R@0.7, and mIoU against the annotated temporal intervals; for the constructed UnAV-100 TVG subset, we use the same metrics on the filtered single-interval test instances described below. For Charades-STA and ActivityNet-Captions, we follow the standard temporal grounding evaluation protocol used by prior video-LLM grounding work. General audio-visual understanding results on WorldSense and Daily-Omni are reported with the official accuracy metric.

\textbf{UnAV-100 TVG Test Subset.}
We additionally construct a temporal video grounding test subset from the UnAV-100~\cite{geng2023dense} test annotations. The original \texttt{test.parquet} contains 5{,}900 audio-visual event annotations over 2{,}167 videos, where each annotation is a tuple of \texttt{video\_id}, event label, and timestamp interval. We first group annotations by \texttt{(video\_id, event\_label)}. Among the 3{,}455 resulting video-event combinations, 1{,}309 correspond to multiple disjoint temporal intervals. Because our evaluation protocol expects one predicted interval per query, these multi-interval cases are ambiguous: a single model output cannot be matched to a unique ground-truth segment without an arbitrary choice. We therefore keep only video-event pairs associated with a single interval, yielding 2{,}146 QA instances over 1{,}610 videos.

We then filter the remaining instances by relative event duration. A large fraction of the single-interval cases nearly spans the whole video: 42.8\% have ground-truth duration ratio above 0.9, and 53.7\% are above 0.7. Such near-full-length events substantially inflate localization metrics, since a trivial predictor that outputs the whole video can already obtain about 0.62 mIoU over the full single-interval pool. To focus the benchmark on genuine temporal localization, we retain only short events whose ground-truth duration divided by video duration is below 0.3. The final UnAV-100 TVG subset contains 653 QA instances over 490 videos. We will release the corresponding JSON file for reproducible evaluation.

For our models and the re-evaluated open-source baselines, we use the same benchmark input format and parse only the final predicted temporal interval. Outputs in seconds are evaluated directly. Outputs in frame indices or discrete temporal slots, when produced by methods whose native protocol uses such representations, are converted back to seconds using the benchmark video duration or frame rate before metric computation, following the corresponding benchmark protocol. If a model produces multiple intervals, we evaluate the first complete interval unless the model's official inference protocol specifies otherwise. Invalid or unparsable outputs, including missing start/end times or intervals outside the video duration after clipping, are counted as failed predictions with zero IoU. For start-end inversions, we reorder the two endpoints before clipping to the valid video range, matching common temporal grounding evaluation practice.

The numbers in Table~\ref{tab:main} and Table~\ref{tab:unav100_tvg} are reported in two ways. For baselines with official benchmark results under the same evaluation setting, we cite the published numbers. For open-source omni-modal models and baselines without directly comparable reported results on AVID, LongVALE, or UnAV-100 TVG, we directly evaluate the released checkpoints; we do not fine-tune these baselines on our reconstructed QA data. The re-evaluated baselines use the same prompt template, video preprocessing budget, timestamp parser, and metric script used for OmniGround. Closed-source Gemini~3.1~Pro is included only as a reference result and is not used for training or model selection. OmniGround and OmniGround$^\dagger$ are the only models in Table~\ref{tab:main} trained on our 580k mixture, whose grounding portion consists of newly reconstructed QA instances derived from the training-split source videos and temporal captions of AVID and LongVALE. Therefore, Table~\ref{tab:main} and Table~\ref{tab:unav100_tvg} should be read as benchmark comparisons under each method's available training protocol; the controlled comparison isolating TACG from data effects is the OmniGround$^\dagger$ row, which uses the same 580k mixture as OmniGround but replaces TACG with standard cross-entropy.

\begin{table}[htbp]
\centering
\setlength{\tabcolsep}{4pt}
\small
\begin{tabular}{l|cc|cc|r}
\toprule
\multirow{2}{*}{Configuration} & \multicolumn{2}{c|}{AVID} & \multicolumn{2}{c|}{LongVALE} & \multirow{2}{*}{$\Delta$\%} \\
& R@0.5 & mIoU & R@0.5 & mIoU & \\
\midrule
Baseline (CE only) & 0.3502 & 0.3153 & 0.1408 & 0.1604 & 0.00 \\
\midrule
$\mathcal{L}_{\text{tg}}$ only & 0.3543 & 0.3223 & \textbf{0.1563} & \underline{0.1722} & \underline{+4.79} \\
$\mathcal{L}_{\text{pg}}$ only & 0.3649 & 0.3282 & 0.1479 & 0.1645 & +3.32 \\
$\mathcal{L}_{\text{tp}}$ only & \underline{0.3671} & \underline{0.3315} & 0.1514 & 0.1663 & +4.41 \\
\textit{Full w/o $\mathcal{L}_{\text{pg}}, \mathcal{L}_{\text{tp}}$} & \textit{0.3558} & \textit{0.3229} & \textit{0.1360} & \textit{0.1563} & \textit{$-$0.08} \\
\textit{Full w/o $\mathcal{L}_{\text{tg}}, \mathcal{L}_{\text{tp}}$} & \textit{0.3350} & \textit{0.3056} & \textit{0.1293} & \textit{0.1488} & \textit{$-$5.16} \\
\textit{Full w/o $\mathcal{L}_{\text{tg}}, \mathcal{L}_{\text{pg}}$} & \textit{0.3534} & \textit{0.3221} & \textit{0.1366} & \textit{0.1560} & \textit{$-$0.29} \\
\midrule
$\mathcal{L}_{\text{tg}} + \mathcal{L}_{\text{pg}}$ & 0.3671 & 0.3301 & 0.1508 & 0.1672 & +4.47 \\
\textit{Full w/o $\mathcal{L}_{\text{tp}}$} & \textit{0.3271} & \textit{0.2989} & \textit{0.1354} & \textit{0.1579} & \textit{$-$3.38} \\
\textit{Full w/o $\mathcal{L}_{\text{pg}}$} & \textit{0.3481} & \textit{0.3150} & \textit{0.1446} & \textit{0.1597} & \textit{$-$0.27} \\
\textit{Full w/o $\mathcal{L}_{\text{tg}}$} & \textit{0.3703} & \textit{0.3315} & \textit{0.1359} & \textit{0.1570} & \textit{+1.51} \\
\midrule
\textbf{Full TACG} & \textbf{0.3814} & \textbf{0.3408} & \underline{0.1546} & \textbf{0.1759} & \textbf{+8.87} \\
\bottomrule
\end{tabular}
\caption{Full component ablation of the TACG consistency loss. Upright rows: addition results with individually optimized weights; \textit{italic rows}: subtraction results inheriting the full-model weights. $\Delta$\%: average relative improvement over the SFT baseline across both benchmarks. \textbf{Bold}/\underline{underline}: best/second best per column.}
\label{tab:edge_ablation_full}
\end{table}

\section{Timestamp Parsing for TACG}
\label{app:tacg_parsing}

All forward-TVG targets follow a standardized template containing a single well-formed start--end timestamp that can be deterministically parsed into digit positions. During training, we locate the timestamp span from the ground-truth answer template and identify the digit-token positions corresponding to the start and end times. The consistency losses are then computed from the logits at these positions, as described in Eq.~\ref{eq:round_method}.

The reconstructed timestamps are converted to seconds and normalized by the video duration before Gaussian projection. We use the timestamp precision provided by the training template and keep the same precision across the start and end fields. The ground-truth distribution is constructed from the annotated interval with the same normalization and Gaussian projection as the prediction, but is detached from the computation graph.


\section{Discretization Weight Sensitivity}
\label{app:discretization_curve}

This section reports the full sweep behind the discretization choice made in the main text. For each of the three ways of turning the per-position digit logits into a predicted timestamp---\emph{Argmax}, \emph{Soft}, and our \emph{Round} (straight-through rounding)---we vary the loss weight $\lambda$ on the predicted-distribution edge and measure mIoU on AVID and LongVALE. Figure~\ref{fig:discretization_bar} summarizes each method at its best $\lambda$ against the cross-entropy (CE) baseline, and Figure~\ref{fig:discretization_curve} gives the complete mIoU-versus-$\lambda$ curves.

The two views agree. Round is the only discretization that stays above the CE baseline on both benchmarks and remains stable across the entire $\lambda$ range, whereas Argmax and Soft dip below the baseline at their best settings and degrade quickly as $\lambda$ grows. This matches the failure modes analyzed in the main text: Argmax's discontinuous digit selection triggers mode-collapse instability, while Soft's non-integer expectation leaves a loss floor that its gradient cannot cross. The robustness of Round across $\lambda$ is what makes it usable without per-benchmark tuning.

\begin{table}[t]
\centering
\setlength{\tabcolsep}{4pt}
\small
\begin{tabular}{l|c}
\toprule
Benchmark (mIoU) & OmniGround (3B) \\
\midrule
AVID & $56.9_{\pm0.4}$ \\
LongVALE & $28.6_{\pm0.5}$ \\
Charades-STA & $38.8_{\pm0.6}$ \\
ActivityNet-Captions & $23.7_{\pm0.4}$ \\
\bottomrule
\end{tabular}
\caption{Multi-seed stability of the 3B final model (mean\,$\pm$\,std over three seeds).}
\label{tab:multiseed_main}
\end{table}

\begin{table}[t]
\centering
\setlength{\tabcolsep}{4pt}
\small
\begin{tabular}{l|cc}
\toprule
Configuration & AVID mIoU & LongVALE mIoU \\
\midrule
Baseline (CE only) & $0.3147_{\pm0.0042}$ & $0.1598_{\pm0.0031}$ \\
$\mathcal{L}_{\text{tg}}$ only & $0.3218_{\pm0.0038}$ & $0.1715_{\pm0.0035}$ \\
$\mathcal{L}_{\text{pg}}$ only & $0.3275_{\pm0.0045}$ & $0.1640_{\pm0.0029}$ \\
$\mathcal{L}_{\text{tp}}$ only & $0.3309_{\pm0.0040}$ & $0.1658_{\pm0.0033}$ \\
$\mathcal{L}_{\text{tg}} + \mathcal{L}_{\text{pg}}$ & $0.3296_{\pm0.0036}$ & $0.1667_{\pm0.0030}$ \\
\textbf{Full TACG} & $\mathbf{0.3401_{\pm0.0034}}$ & $\mathbf{0.1753_{\pm0.0028}}$ \\
\bottomrule
\end{tabular}
\caption{Multi-seed stability of the component ablation (cf.\ Table~\ref{tab:edge_ablation}), mean\,$\pm$\,std over three seeds.}
\label{tab:multiseed_ablation}
\end{table}

\section{Weight Ablation}
\label{app:weight_ablation}

We ablate the loss weight $\lambda$ of each edge by sweeping across a range of values with both JS divergence and Wasserstein-1 (W1) as distance metrics.

As shown in Figure~\ref{fig:weight_sensitivity}, all edges share the same fundamental trade-off: when $\lambda$ is too small, the supervision signal is too weak to meaningfully guide learning; when too large, the distribution-matching gradient overwhelms the cross-entropy loss and degrades generation quality. The optimal weight and distance metric, however, differ across edges. The ground-truth-anchored edges admit stronger supervision: $\mathcal{L}_{\text{pg}}$ peaks with W1 at $\lambda_{\text{pg}}=0.2$, whose distance-proportional gradients give a smoother landscape that tolerates large weights, whereas $\mathcal{L}_{\text{tg}}$ favors JS at $\lambda_{\text{tg}}=0.1$ with a broad effective range---for noisy attention supervision, JS measures distributional overlap without sensitivity to the absolute magnitude of misalignment and is thus more stable. The self-supervised edge $\mathcal{L}_{\text{tp}}$ exhibits a narrower effective range, as it enforces consistency between two model-internal signals without external supervision and excessively large weights cause the distributions to reinforce each other's errors; W1 with $\lambda_{\text{tp}}=0.1$ achieves the best balance. Beyond the per-edge sweeps, Figure~\ref{fig:tgpg_surface} maps the joint $\mathcal{L}_{\text{tg}}{+}\mathcal{L}_{\text{pg}}$ response surface over a $2$D weight grid, confirming that the two ground-truth-anchored edges produce a broad region of improvement over the SFT baseline rather than a brittle optimum.

\section{Full Component Ablation}
\label{app:edge_ablation_full}

Table~\ref{tab:edge_ablation_full} reports the complete component ablation, including both addition results (upright rows, with individually optimized weights via progressive search) and subtraction results (\textit{italic rows}, which inherit weights from the full configuration); the main text (Table~\ref{tab:edge_ablation}) shows only the addition rows.

Two observations stand out. First, among all single-edge removals from the full model, dropping the self-supervised attention--prediction edge $\mathcal{L}_{\text{tp}}$ causes the steepest degradation ($-3.38\%$), reinforcing the main-text conclusion that it is the key contributor rather than a redundant term. Second, the subtraction rows---which reuse the full model's weights instead of re-optimizing for the smaller edge subset---frequently under-perform their addition counterparts, and can even fall below the baseline. This shows that the edge weights do not transfer across subsets, and is why we report both the addition and subtraction views rather than a single ablation direction.

\section{Multi-Seed Stability}
\label{app:multiseed}

The results in the main text come from a single run with a fixed seed (42). To verify that the reported margins are not seed artifacts---particularly the small full-scale gaps and the closely spaced single-edge ablations---we repeat the key experiments over three seeds and report the mean and standard deviation. Table~\ref{tab:multiseed_main} covers the 3B final model on the four grounding benchmarks, and Table~\ref{tab:multiseed_ablation} repeats the component ablation of Table~\ref{tab:edge_ablation}.
Across both tables, the mean stays close to the single-seed estimate and the ordering of configurations is preserved, indicating that the reported gains are stable rather than seed-specific.

%% file: references.bib
@inproceedings{2dtan,
  title={Learning 2D Temporal Adjacent Networks for Moment Localization with Natural Language},
  author={Zhang, Songyang and Peng, Houwen and Fu, Jianlong and Luo, Jiebo},
  booktitle={AAAI},
  year={2020}
}

@inproceedings{univtg,
  title={UniVTG: Towards Unified Video-Language Temporal Grounding},
  author={Lin, Kevin Qinghong and Zhang, Pengchuan and Chen, Joya and Pramanick, Shraman and Gao, Difei and Wang, Alex Jinpeng and Yan, Rui and Shou, Mike Zheng},
  booktitle={Proceedings of the IEEE/CVF International Conference on Computer Vision (ICCV)},
  pages={2794--2804},
  year={2023}
}

@inproceedings{snag,
  title={SnAG: Scalable and Accurate Video Grounding},
  author={Mu, Fangzhou and Mo, Sicheng and Li, Yin},
  booktitle={Proceedings of the IEEE/CVF Conference on Computer Vision and Pattern Recognition (CVPR)},
  pages={18930--18940},
  year={2024}
}

@inproceedings{eatr,
  title={Knowing Where to Focus: Event-aware Transformer for Video Grounding},
  author={Jang, Jinhyun and Park, Jungin and Kim, Jin and Kwon, Hyeongjun and Sohn, Kwanghoon},
  booktitle={Proceedings of the IEEE/CVF International Conference on Computer Vision (ICCV)},
  pages={13846--13856},
  year={2023}
}

@inproceedings{timechat,
  title={{TimeChat}: A Time-sensitive Multimodal Large Language Model for Long Video Understanding},
  author={Ren, Shuhuai and Yao, Linli and Li, Shicheng and Sun, Xu and Hou, Lu},
  booktitle={CVPR},
  year={2024}
}

@article{hawkeye,
  title={{HawkEye}: Training Video-Text {LLMs} for Grounding Text in Videos},
  author={Wang, Yueqian and Meng, Xiaojun and Liang, Jianxin and Wang, Yuxuan and Liu, Qun and Zhao, Dongyan},
  journal={arXiv preprint arXiv:2403.10228},
  year={2024}
}

@inproceedings{vtimellm,
  title={{VTimeLLM}: Empower {LLM} to Grasp Video Moments},
  author={Huang, Bin and Wang, Xin and Chen, Hong and Song, Zihan and Zhu, Wenwu},
  booktitle={CVPR},
  year={2024}
}

@inproceedings{timesuite,
  title={{TimeSuite}: Improving {MLLMs} for Long Video Understanding via Grounded Tuning},
  author={Zeng, Xiangyu and Li, Kunchang and Wang, Chenting and Li, Xinhao and Jiang, Tianxiang and Yan, Ziang and Li, Songze and Shi, Yansong and Yue, Zhengrong and Wang, Yi and Wang, Yali and Qiao, Yu and Wang, Limin},
  booktitle={ICLR},
  year={2025}
}

@inproceedings{videochatflash,
  title={VideoChat-Flash: Hierarchical Compression for Long-Context Video Modeling},
  author={Li, Xinhao and Wang, Yi and Yu, Jiashuo and Zeng, Xiangyu and Zhu, Yuhan and Huang, Haian and Gao, Jianfei and Li, Kunchang and He, Yinan and Wang, Chenting and Qiao, Yu and Wang, Yali and Wang, Limin},
  booktitle={The Fourteenth International Conference on Learning Representations (ICLR)},
  year={2026},
  url={https://openreview.net/forum?id=MUjdNcfNPv}
}

@inproceedings{trace,
  title={{TRACE}: Temporal Grounding Video {LLM} via Causal Event Modeling},
  author={Guo, Yongxin and Liu, Jingyu and Li, Mingda and Liu, Qingbin and Chen, Xi and Tang, Xiaoying},
  booktitle={The Thirteenth International Conference on Learning Representations (ICLR)},
  year={2025},
  url={https://openreview.net/forum?id=14fFV0chUS}
}

@inproceedings{chatvtg,
  title={{ChatVTG}: Video Temporal Grounding via Chat with Video Dialogue Large Language Models},
  author={Qu, Mengxue and Chen, Xiaodong and Liu, Wu and Li, Alicia and Zhao, Yao},
  booktitle={Proceedings of the IEEE/CVF Conference on Computer Vision and Pattern Recognition Workshops (CVPRW)},
  pages={1847--1856},
  year={2024}
}

@inproceedings{timer1,
  title={Time-{R1}: Post-Training Large Vision Language Model for Temporal Video Grounding},
  author={Wang, Ye and Wang, Ziheng and Xu, Boshen and Du, Yang and Lin, Kejun and Xiao, Zihan and Yue, Zihao and Ju, Jianzhong and Zhang, Liang and Yang, Dingyi and Fang, Xiangnan and He, Zewen and Luo, Zhenbo and Wang, Wenxuan and Lin, Junqi and Luan, Jian and Jin, Qin},
  booktitle={NeurIPS},
  year={2025}
}

@article{chronusomni,
  title={{ChronusOmni}: Improving Time Awareness of Omni Large Language Models},
  author={Chen, Yijing and Wu, Yihan and Guan, Kaisi and Ren, Yuchen and Wang, Yuyue and Song, Ruihua and Ru, Liyun},
  journal={arXiv preprint arXiv:2512.09841},
  year={2025}
}

@inproceedings{edvtg,
  title={Enrich and Detect: Video Temporal Grounding with Multimodal {LLMs}},
  author={Pramanick, Shraman and Mavroudi, Effrosyni and Song, Yale and Chellappa, Rama and Torresani, Lorenzo and Afouras, Triantafyllos},
  booktitle={ICCV},
  year={2025}
}

@inproceedings{groundedvideollm,
  title={Grounded-{VideoLLM}: Sharpening Fine-grained Temporal Grounding in Video Large Language Models},
  author={Wang, Haibo and Xu, Zhiyang and Cheng, Yu and Diao, Shizhe and Zhou, Yufan and Cao, Yixin and Wang, Qifan and Ge, Weifeng and Huang, Lifu},
  booktitle={EMNLP Findings},
  year={2025}
}

@inproceedings{vtgllm,
  title={{VTG-LLM}: Integrating Timestamp Knowledge into Video {LLMs} for Enhanced Video Temporal Grounding},
  author={Guo, Yongxin and Liu, Jingyu and Li, Mingda and Cheng, Dingxin and Tang, Xiaoying and Sui, Dianbo and Liu, Qingbin and Chen, Xi and Zhao, Kevin},
  booktitle={AAAI},
  year={2025}
}

@inproceedings{numpro,
  title={Number it: Temporal Grounding Videos like Flipping Manga},
  author={Wu, Yongliang and Hu, Xinting and Sun, Yuyang and Zhou, Yizhou and Zhu, Wenbo and Rao, Fengyun and Schiele, Bernt and Yang, Xu},
  booktitle={CVPR},
  year={2025}
}

@inproceedings{revisionllm,
  title={{ReVisionLLM}: Recursive Vision-Language Model for Temporal Grounding in Hour-Long Videos},
  author={Hannan, Tanveer and Islam, Md Mohaiminul and Gu, Jindong and Seidl, Thomas and Bertasius, Gedas},
  booktitle={Proceedings of the IEEE/CVF Conference on Computer Vision and Pattern Recognition (CVPR)},
  pages={19012--19022},
  year={2025}
}

@article{videochat,
  title={VideoChat: Chat-Centric Video Understanding},
  author={Li, Kunchang and He, Yinan and Wang, Yi and Li, Yizhuo and Wang, Wenhai and Luo, Ping and Wang, Yali and Wang, Limin and Qiao, Yu},
  journal={arXiv preprint arXiv:2305.06355},
  year={2023}
}

@inproceedings{videochatgpt,
  title={Video-ChatGPT: Towards Detailed Video Understanding via Large Vision and Language Models},
  author={Maaz, Muhammad and Rasheed, Hanoona and Khan, Salman and Khan, Fahad},
  booktitle={Proceedings of the 62nd Annual Meeting of the Association for Computational Linguistics (Volume 1: Long Papers)},
  pages={12585--12602},
  publisher={Association for Computational Linguistics},
  year={2024},
  doi={10.18653/v1/2024.acl-long.679},
  url={https://aclanthology.org/2024.acl-long.679/}
}

@inproceedings{videollama,
  title={Video-LLaMA: An Instruction-tuned Audio-Visual Language Model for Video Understanding},
  author={Zhang, Hang and Li, Xin and Bing, Lidong,Zesen and Leng, Sicong and Zhang, Hang and Xin, Yifei and Li},
  booktitle={Proceedings of the 2023 Conference on Empirical Methods in Natural Language Processing: System Demonstrations},
  pages={543--553},
  publisher={Association for Computational Linguistics},
  year={2023},
  doi={10.18653/v1/2023.emnlp-demo.49},
  url={https://aclanthology.org/2023.emnlp-demo.49/}
}

@article{videollama2,
  title={{VideoLLaMA} 2: Advancing Spatial-Temporal Modeling and Audio Understanding in Video-{LLMs}},
  author={Cheng, Zesen and Leng, Sicong and Zhang, Hang and Xin, Yifei and Li, Xin and Chen, Guanzheng and Zhu, Yongxin and Zhang, Wenqi and Luo, Ziyang and Zhao, Deli and Bing, Lidong},
  journal={arXiv preprint arXiv:2406.07476},
  year={2024}
}

@inproceedings{pandagpt,
  title={PandaGPT: One Model To Instruction-Follow Them All},
  author={Su, Yixuan and Lan, Tian and Li, Huayang and Xu, Jialu and Wang, Yan and Cai, Deng},
  booktitle={Proceedings of the 1st Workshop on Taming Large Language Models: Controllability in the Era of Interactive Assistants!},
  pages={11--23},
  publisher={Association for Computational Linguistics},
  year={2023},
  url={https://aclanthology.org/2023.tllm-1.2/}
}

@inproceedings{nextgpt,
  title={NExT-GPT: Any-to-Any Multimodal LLM},
  author={Wu, Shengqiong and Fei, Hao and Qu, Leigang and Ji, Wei and Chua, Tat-Seng},
  booktitle={Proceedings of the 41st International Conference on Machine Learning},
  pages={53366--53397},
  volume={235},
  series={Proceedings of Machine Learning Research},
  publisher={PMLR},
  year={2024},
  url={https://proceedings.mlr.press/v235/wu24e.html}
}

@article{ola,
  title={Ola: Pushing the Frontiers of Omni-Modal Language Model},
  author={Liu, Zuyan and Dong, Yuhao and Wang, Jiahui and Liu, Ziwei and Hu, Winston and Lu, Jiwen and Rao, Yongming},
  journal={arXiv preprint arXiv:2502.04328},
  year={2025}
}

@article{qwen25omni,
  title={Qwen2.5-Omni Technical Report},
  author={Xu, Jin and Guo, Zhifang and He, Jinzheng and Hu, Hangrui and He, Ting and Bai, Shuai and Chen, Keqin and Wang, Jialin and Fan, Yang and Dang, Kai and Zhang, Bin and Wang, Xiong and Chu, Yunfei and Lin, Junyang},
  journal={arXiv preprint arXiv:2503.20215},
  year={2025}
}

@article{qwen3omni,
  title={Qwen3-Omni Technical Report},
  author={Xu, Jin and Guo, Zhifang and Hu, Hangrui and Chu, Yunfei and Wang, Xiong and He, Jinzheng and Wang, Yuxuan and Shi, Xian and He, Ting and Zhu, Xinfa and Lv, Yuanjun and Wang, Yongqi and Guo, Dake and Wang, He and Ma, Linhan and Zhang, Pei and Zhang, Xinyu and Hao, Hongkun and Guo, Zishan and Yang, Baosong and Zhang, Bin and Ma, Ziyang and Wei, Xipin and Bai, Shuai and Chen, Keqin and Liu, Xuejing and Wang, Peng and Yang, Mingkun and Liu, Dayiheng and Ren, Xingzhang and Zheng, Bo and Men, Rui and Zhou, Fan and Yu, Bowen and Yang, Jianxin and Yu, Le and Zhou, Jingren and Lin, Junyang},
  journal={arXiv preprint arXiv:2509.17765},
  year={2025}
}

@misc{gemini,
  title={Gemini 3.1 Pro Model Card},
  author={{Google DeepMind}},
  year={2026},
  month={February},
  howpublished={Model card},
  url={https://deepmind.google/models/model-cards/gemini-3-1-pro/}
}

@inproceedings{longvale,
  title={{LongVALE}: Vision-Audio-Language-Event Benchmark Towards Time-Aware Omni-Modal Perception of Long Videos},
  author={Geng, Tiantian and Zhang, Jinrui and Wang, Qingni and Wang, Teng and Duan, Jinming and Zheng, Feng},
  booktitle={CVPR},
  year={2025}
}

@article{avid,
  title={{AVID}: A Benchmark for Omni-Modal Audio-Visual Inconsistency Understanding via Agent-Driven Construction},
  author={Chen, Zixuan and Wang, Depeng and Lin, Hao and Luo, Li and Xu, Ke and Guo, Ya and Zhu, Huijia and Sun, Tanfeng and Jiang, Xinghao},
  journal={arXiv preprint arXiv:2604.13593},
  year={2026},
  url={https://arxiv.org/abs/2604.13593}
}

@inproceedings{worldsense,
  title={{WorldSense}: Evaluating Real-world Omnimodal Understanding for Multimodal {LLMs}},
  author={Jack Hong and Shilin Yan and Jiayin Cai and Xiaolong Jiang and Yao Hu and Weidi Xie},
  booktitle={International Conference on Learning Representations (ICLR)},
  year={2026},
  note={arXiv preprint arXiv:2502.04326},
  url={https://arxiv.org/abs/2502.04326}
}

@article{dailyomni,
  title={Daily-Omni: Towards Audio-Visual Reasoning with Temporal Alignment across Modalities},
  author={Zhou, Ziwei and Wang, Rui and Wu, Zuxuan and Jiang, Yu-Gang},
  journal={arXiv preprint arXiv:2505.17862},
  year={2025}
}

@inproceedings{avqa,
  title={{AVQA}: A Dataset for Audio-Visual Question Answering on Videos},
  author={Yang, Pinci and Wang, Xin and Duan, Xuguang and Chen, Hong and Hou, Runze and Jin, Cong and Zhu, Wenwu},
  booktitle={ACM MM},
  year={2022}
}

@article{llavavideo,
  title={Video Instruction Tuning with Synthetic Data},
  author={Zhang, Yuanhan and Wu, Jinming and Li, Wei and Li, Bo and Ma, Zejun and Liu, Ziwei and Li, Chunyuan},
  journal={arXiv preprint arXiv:2410.02713},
  year={2024}
}

@article{fortisavqa,
  title={{FortisAVQA} and {MAVEN}: A Benchmark Dataset and Debiasing Framework for Robust Multimodal Reasoning},
  author={Ma, Jie and Gao, Zhitao and Chai, Qi and Liu, Jun and Wang, Pinghui and Tao, Jing and Su, Zhou},
  journal={arXiv preprint arXiv:2504.00487},
  year={2025}
}

@article{videoopd,
  title={Video-{OPD}: Efficient Post-Training of Multimodal Large Language Models for Temporal Video Grounding via On-Policy Distillation},
  author={Li, Jiaze and Yin, Hao and Xu, Haoran and Xu, Boshen and Tan, Wenhui and He, Zewen and Ju, Jianzhong and Luo, Zhenbo and Luan, Jian},
  journal={arXiv preprint arXiv:2602.02994},
  year={2026}
}

@article{zhao2024first,
  title={The First to Know: How Token Distributions Reveal Hidden Knowledge in Large Vision-Language Models?},
  author={Zhao, Qinyu and Xu, Ming and Gupta, Kartik and Asthana, Akshay and Zheng, Liang and Gould, Stephen},
  journal={arXiv preprint arXiv:2403.09037},
  year={2024}
}

@inproceedings{Palit_2023_ICCV,
  author={Palit, Vedant and Pandey, Rohan and Arora, Aryaman and Liang, Paul Pu},
  title={Towards Vision-Language Mechanistic Interpretability: A Causal Tracing Tool for {BLIP}},
  booktitle={Proceedings of the IEEE/CVF International Conference on Computer Vision (ICCV) Workshops},
  month={October},
  year={2023},
  pages={2856--2861}
}

@inproceedings{activitynet,
  title={{ActivityNet}: A Large-Scale Video Benchmark for Human Activity Understanding},
  author={Caba Heilbron, Fabian and Escorcia, Victor and Ghanem, Bernard and Carlos Niebles, Juan},
  booktitle={CVPR},
  year={2015}
}

@inproceedings{charades,
  title={{TALL}: Temporal Activity Localization via Language Query},
  author={Gao, Jiyang and Sun, Chen and Yang, Zhenheng and Nevatia, Ram},
  booktitle={ICCV},
  year={2017},
  note={Charades-STA benchmark}
}

@inproceedings{geng2023dense,
  title={Dense-Localizing Audio-Visual Events in Untrimmed Videos: A Large-Scale Benchmark and Baseline},
  author={Geng, Tiantian and Wang, Teng and Duan, Jinming and Cong, Runmin and Zheng, Feng},
  booktitle={Proceedings of the IEEE/CVF Conference on Computer Vision and Pattern Recognition},
  pages={22942--22951},
  year={2023}
}
